\documentclass{article} % For LaTeX2e
\usepackage{iclr2025_conference,times}

\usepackage{amsmath,amsfonts,bm}

\def\eqref#1{equation~\ref{#1}}
\def\1{\bm{1}}

\DeclareMathAlphabet{\mathsfit}{\encodingdefault}{\sfdefault}{m}{sl}
\SetMathAlphabet{\mathsfit}{bold}{\encodingdefault}{\sfdefault}{bx}{n}

\newcommand{\R}{\mathbb{R}}

\usepackage{hyperref}
\usepackage{url}
\usepackage{makecell}

\usepackage{xspace}
\newcommand{\NAME}{WeatherODE\xspace}

\usepackage{booktabs}
\usepackage{graphicx}
\usepackage{multicol, multirow, xcolor, threeparttable}

\usepackage{pifont}
\usepackage{afterpage}
\newcommand{\correct}{\text{\ding{52}}}
\newcommand{\wrong}{\text{\ding{56}}}

\usepackage[capitalize]{cleveref}
\crefname{section}{Section}{Sections}
\Crefname{section}{Section}{Sections}
\Crefname{table}{Table}{Tables}
\Crefname{figure}{Figure}{Figures}
\crefformat{equation}{Equation ~#2#1#3}

\definecolor{bl}{rgb}{0.25, 0.5, 0.9}

\title{
Mitigating Time Discretization Challenges with WeatherODE: A Sandwich Physics-Driven Neural ODE for Weather Forecasting
}

\iclrfinalcopy

\author{Peiyuan Liu$^{1,}$\thanks{Equal contribution}, Tian Zhou$^{2,}$\footnotemark[1], Liang Sun$^{2}$, Rong Jin$^{2}$
\\
$^{1}$Tsinghua University, $^{2}$Alibaba Group\\
 \texttt{lpy23@mails.tsinghua.edu.cn}, \
 \texttt{tian.zt@alibaba-inc.com},\\
 \texttt{liang.sun@alibaba-inc.com}, \
 \texttt{rongjinemail@gmail.com}
}

\begin{document}

\maketitle

\begin{abstract}
In the field of weather forecasting, traditional models often grapple with discretization errors and time-dependent source discrepancies, which limit their predictive performance. In this paper, we present WeatherODE, a novel one-stage, physics-driven ordinary differential equation (ODE) model designed to enhance weather forecasting accuracy. By leveraging wave equation theory and integrating a time-dependent source model, WeatherODE effectively addresses the challenges associated with time-discretization error and dynamic atmospheric processes. Moreover, we design a CNN-ViT-CNN sandwich structure, facilitating efficient learning dynamics tailored for distinct yet interrelated tasks with varying optimization biases in advection equation estimation. Through rigorous experiments, WeatherODE demonstrates superior performance in both global and regional weather forecasting tasks, outperforming recent state-of-the-art approaches by significant margins of over 40.0\% and 31.8\% in root mean square error (RMSE), respectively. The source code is available at \url{https://github.com/DAMO-DI-ML/WeatherODE}.
\end{abstract}

\section{Introduction}

Weather forecasting is a cornerstone of modern society, affecting key industries like agriculture, transportation, and disaster management~\citep{coiffier-fundamentals_2011}. Accurate predictions help mitigate the effects of extreme weather and optimize economic operations. Recent advancements in high-performance computing have significantly boosted the accuracy and speed of numerical weather forecasting (NWP)~\citep{bauer2015quiet, nwp1, nwp2}.

The swift advancement of deep learning has opened up a promising avenue for weather forecasting~\citep{weyn2019can,scher2019weather,rasp2020weatherbench,weyn2021sub,pangu,fourcastnet,hu2023swinvrnn}. However, the existing weather forecasting models based on deep learning often fail to fully account for the key physical mechanisms governing small-scale, complex nonlinear atmospheric phenomena, such as turbulence, convection, and airflow. These dynamic processes are crucial to the formation and evolution of weather systems, but most models focus on learning statistical correlations from historical data instead of explicitly extracting or integrating these physical dynamics.
Furthermore, these models typically rely on fixed time intervals (e.g., every 6 hours) for predictions, limiting their applicability to 
varying temporal scales. Consequently, separate models are often required for different forecast periods~\citep{pangu}, which constrains flexibility and reduces generalization.

\begin{figure}[!t]
    \centering
    \includegraphics[width=\linewidth]{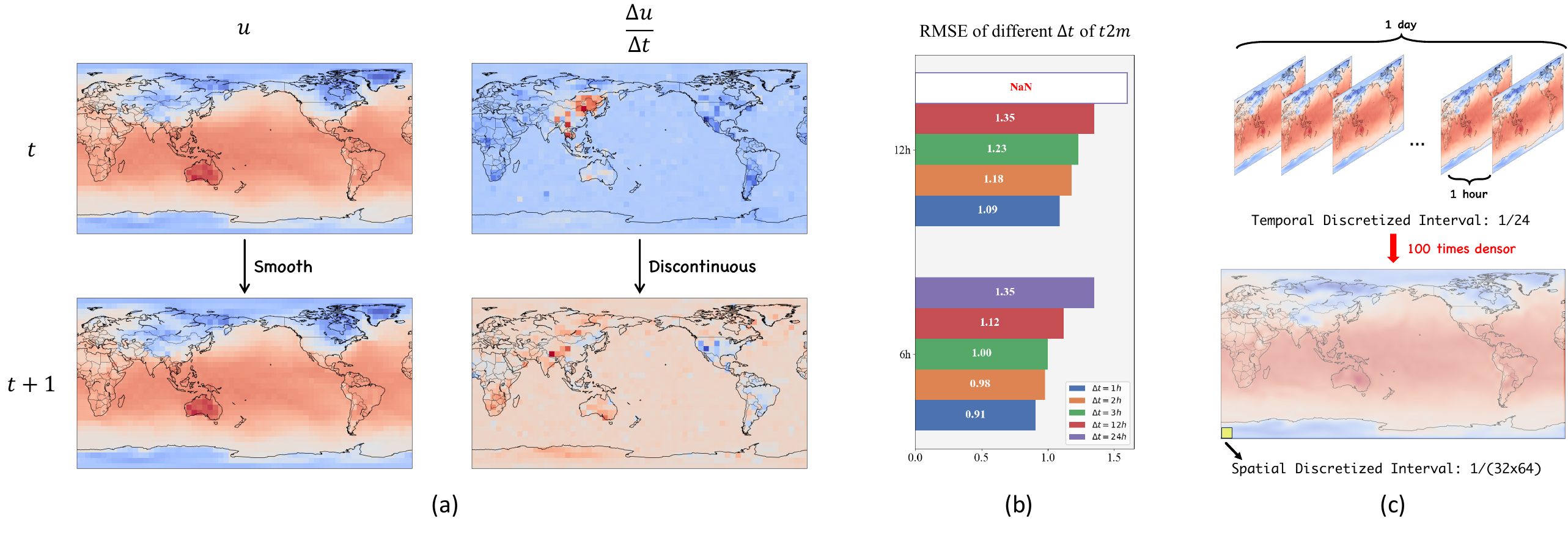}
    \caption{(a) Comparison of two-meter temperature (\( t2m \)) and its discrete-time derivative over a 1-hour interval. While the temperature evolves continuously, the discrete-time derivative exhibits discontinuities, leading to discretization errors. (b) Latitude-weighted RMSE for \( t2m \) using models trained with different time intervals (\(\Delta t\)) for estimating initial velocity. Larger \(\Delta t\) values result in worse performance and can even lead to numerical instability (NaN). See \cref{tab:ablation_time_interval} for full results. (c) Comparison of temporal and spatial discretization intervals in the 5.625° ERA5 dataset. The spatial discretization is 100 times denser than the temporal discretization.}
    \label{fig:intro}
\end{figure}

Another line of research utilizes neural ODEs~\citep{neuralODE} that incorporate partial differential equations to guide the physical dynamics of weather forecasting. Among these methods, the advection continuity equation stands out as a key equation governing many weather indicators:
\begin{align} \label{eq:ce}
\frac{\partial u}{\partial t} + \:\: \underbrace{v \cdot \nabla u \: + u \nabla \cdot v}_{\text{Advection}} = \underbrace{s}_{\text{Source}},
\end{align}
where \( u \) represents a atmospheric variable evolving over space and time, driven by the flow velocity \( v \) and the source term \( s \). A recent study, ClimODE~\citep{climode}, effectively employs this equation and achieves state-of-the-art performance. %However, there are several inherent challenges when solving such equations using neural ODEs: (1) Estimating the initial velocity is necessary. Typically, we use time discretization to estimate the gradient (velocity); however, we face a constraint due to a 1-hour discretization limit imposed by the temporal resolution of the ERA5 dataset, which serves as the default training data for most global weather forecasting models. As shown in \cref{fig:intro}a, it is evident that velocity estimation is far from continuous, despite the observed variable being relatively smooth and continuous. Furthermore, we demonstrate in \cref{fig:intro}b that using larger discretization intervals for velocity estimation would significantly further hinder our forecasting performance. This indicates that 1-hour estimates can introduce significant errors. Coarse calculations from 5.625° ERA5 data \citep{weatherbench} reveal a temporal resolution of 1/24 and a spatial resolution of 1/(32×64), resulting in the spatial domain nearly 100 times denser, which could reduce errors from temporal discretization (\cref{fig:intro}c). (2) To solve the advection equation, three key components must be addressed: initial velocity estimation, solving the advection equation itself, and accounting for the error term that arises from deviations in reality. These three tasks require different types of interaction. Global and long-term interactions will govern the advection process, while local and short-term interactions will dictate the velocity estimation and equation's overall deviations. (3) The source term should be modeled as time-dependent.
However, there are several inherent challenges when solving such equations using neural ODEs. 
Firstly, the accurate estimation of the initial velocity is crucial to the weather forecasting performance. Unfortunately, current methods typically rely on time discretization to estimate atmospheric time gradients for velocity calculation and cannot achieve satisfying accuracy. In particular, we face a constraint due to a 1-hour discretization limit imposed by the temporal resolution of the ERA5 dataset, which is usually chosen for training of deep models including most global weather forecasting models. As shown in \cref{fig:intro}a, it is evident that velocity estimation is far from continuous, despite the observed variable being relatively smooth and continuous. Furthermore, we demonstrate in \cref{fig:intro}b that using larger discretization intervals for velocity estimation would significantly hinder our forecasting performance. This indicates that 1-hour estimates can introduce significant errors. On the other hand, we note that coarse calculations from 5.625° ERA5 data~\citep{weatherbench} reveal a temporal resolution of 1/24 and a spatial resolution of 1/(32×64), resulting in the spatial domain nearly 100 times denser, which can help to reduce errors from temporal discretization (\cref{fig:intro}c).
Secondly, to better solve the advection equation, we need to consider three key components carefully, including the initial velocity estimation, solving the advection equation itself, and the error term arising from deviations in reality. Due to their physical nature, they call for different modeling. For example, global and long-term interactions govern the advection process, while local and short-term interactions dictate the velocity estimation and equation's overall deviations.
Lastly, the source term should be modeled as time-dependent for better estimation.

% \vskip -0.1in

To address these challenges, we propose \NAME, a one-stage, physics-driven ODE model for weather forecasting. It leverages the wave equation, widely used in atmospheric simulations, to improve the estimation of initial velocity using more precise spatial information \( \nabla u \). This approach reduces the time-discretization errors introduced by using \( \frac{\Delta u}{\Delta t} \). Additionally, we introduce a time-dependent source model that effectively captures the evolving dynamics of the source term. Furthermore, we have meticulously crafted the model architecture to seamlessly integrate local feature extraction with global context modeling, promoting efficient learning dynamics tailored for three tasks in advection equation estimation. Our contributions can be summarized as follows:
\begin{itemize}
    \item We conduct thorough experiments to identify and demonstrate the issues of discretization error and time-dependent source error, both of which significantly hinder the performance of current physics-informed weather forecasting models.
    
    \item We propose \NAME, a one-stage, physics-driven ODE model for weather forecasting that utilizes wave equation theory and a time-dependent source model to address the identified challenges. To solve the advection equation more accurately, we conduct a comprehensive analysis of the architectural design of the CNN-ViT-CNN sandwich structure, facilitating efficient learning dynamics tailored for distinct yet interrelated tasks with varying optimization biases. % in advection equation estimation

    \item \NAME demonstrates impressive performance in both global and regional weather forecasting tasks, significantly surpassing the recent state-of-the-art methods by margins of 40.0\% and 31.8\% in RMSE, respectively.

    %enhance the estimation of the initial velocity and integrate temporal dynamics into the source estimation. 
\end{itemize}

\section{Related Works}

The most advanced weather forecasting techniques predominantly rely on Numerical Weather Prediction (NWP)~\citep{nwp1, nwp2}, which employs a set of equations solved on supercomputers to model and predict the atmosphere. 
% translates atmospheric physical equations into computational algorithms that run on supercomputers. 
While NWP has achieved promising results, it is resource-intensive, requiring significant computational power and domain expertise to define the appropriate physical equations.

Deep learning-based weather forecasting adopts a data-driven approach to learning the spatio-temporal relationships between atmospheric variables. These methods can be broadly classified into Graph Neural Networks (GNN) and Transformer-based methods. GNN-based methods~\citep{graphcast, keisler} treat the Earth as a graph and use graph neural networks to predict weather patterns. Transformer-based approaches have shown significant success in weather forecasting due to their scalability~\citep{fuxi, fengwu, fengwu-ghr, attention}. 
For example, Pangu~\citep{pangu} employs a 3D Swin Transformer~\citep{swintransformer} and an autoregressive model to accelerate inference. 
Fengwu~\citep{fengwu} models atmospheric variables as separate modalities and uses a replay buffer for optimization, with Fengwu-GHR~\citep{fengwu-ghr} subsequently extending the approach to higher-resolution data.  Additionally, ClimaX~\citep{climax} and Aurora~\citep{aurora} introduce a pretraining-finetuning framework, where models are first pretrained on physics-simulated data and then finetuned on real-world data. However, these models frequently neglect the fundamental physical dynamics of the atmosphere and are limited to providing fixed lead time for each prediction.

Physics-driven methods, which integrate physical constraints in the form of partial differential equations (PDEs)~\citep{pde} into neural networks, have gained increasing attention in recent years~\citep{PINN, PINO}. In weather forecasting, DeepPhysiNet~\citep{deepphysinet} incorporates physical laws into the loss function, marking an initial attempt to combine neural networks with PDEs. ClimODE~\citep{climode} advances further by leveraging the continuity equation to express the weather forecasting process as a full PDE system modeled using neural ODEs~\citep{neuralODE}. NeuralGCM~\citep{neuralGCM} incorporates more physical constraints and designs neural networks to function as a dynamic core. However, it is complex and difficult to modify, as it operates with over a dozen ODE functions similar to the NWP method. In contrast, our proposed WeatherODE offers a more straightforward and efficient foundation for ongoing improvements.

% However, these works still focus on fixed lead-time predictions and do not address the optimization challenges of integrating neural networks with PDEs.

\section{Method}

In this section, we first introduce the overall ODE modeling framework for weather forecasting in \cref{sec:ode_framework}. We then describe the specific designs of the Velocity Model, Advection ODE, and Source Model in \cref{sec:velocity_model}, \cref{sec:advection_ode}, and \cref{sec:source_model}, respectively. We present the overarching design choices for our CNN-ViT-CNN sandwich structure in \cref{sec:sandwich}. Finally, we end up with the multi-task learning strategy in \cref{sec:multi_task_learning}.

\begin{figure}[htb]
    \centering
    \includegraphics[width=\linewidth]{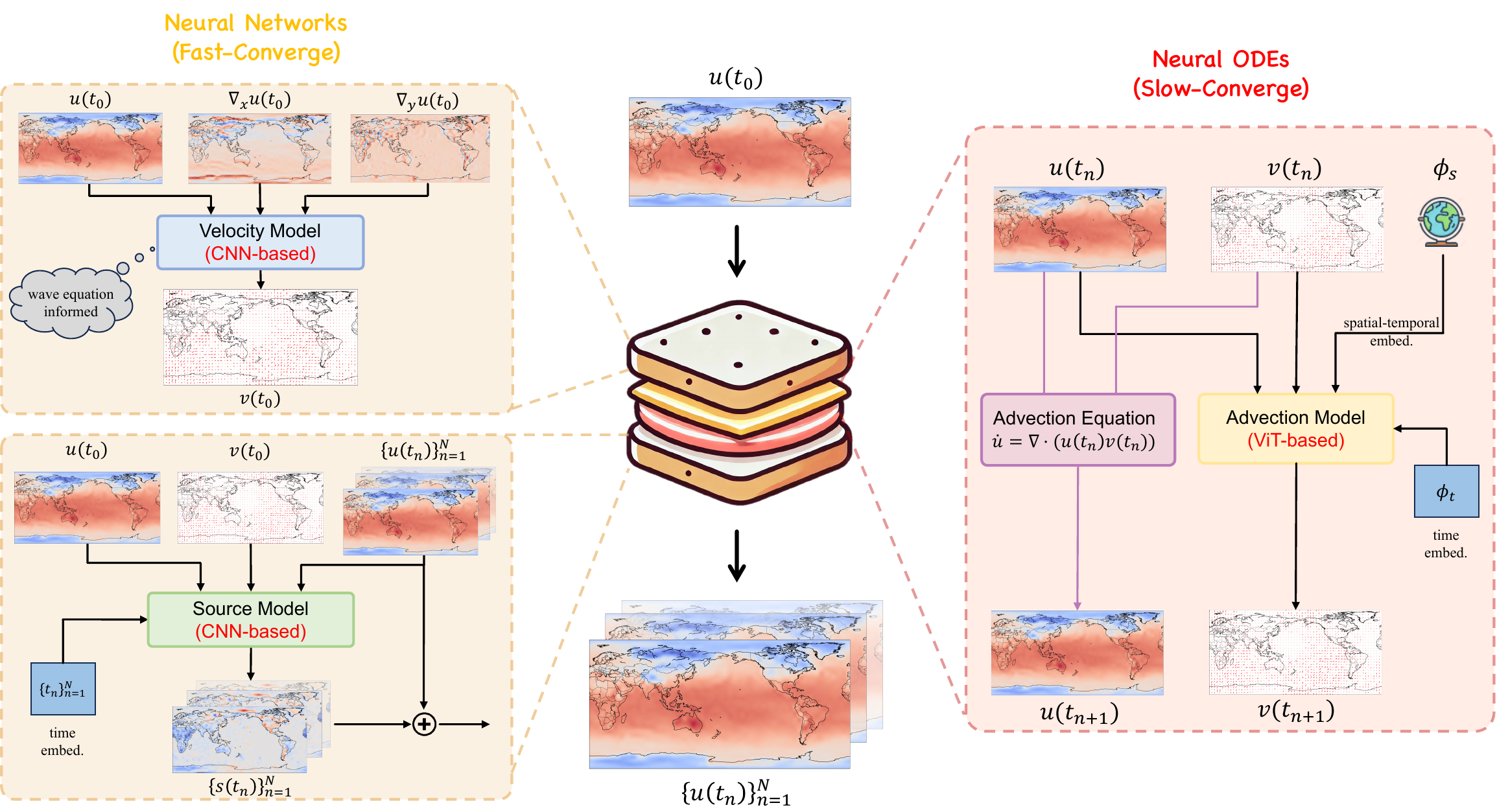}
    \caption{\small Overall architecture of \NAME. \NAME adopts a sandwich-like structure for atmosphere modeling. The top and bottom parts use fast-converging neural networks (CNN-based) to estimate the initial velocity and source term, while the central layer employs a slower-converging neural ODE (ViT-based) to model the atmospheric advection process. This design ensures stability when training the neural ODE to solve the numerical solution. More analyses are in \cref{sec:sandwich} and \cref{sec:abla_optimize}.}
    \label{fig:pipeline}
    \vskip -0.2in
\end{figure}

\subsection{ODE Framework for Weather Dynamics}
\label{sec:ode_framework}

We can model the atmosphere as a spatio-temporal process \( \mathbf{u}(x, y, t) = (u_1(x, y, t), \ldots, u_K(x, y, t)) \in \R^K \), where \( K \) represents the number of distinct atmospheric variables \( u_k(x, y, t) \in \R \), evolving over continuous time \( t \) and spatial coordinates \( (x, y) \in [0, H] \times [0, W] \), $H$ and $W$ are the height and width, respectively. Each quantity or atmospheric variable is driven by a velocity field \( v_k(x,y,t) \in \R^{2K}\) and influenced by a source term \( s_k(x,y,t) \in \R^K\). For simplicity, we first omit the index \( k \) since all quantities are treated equally, and then drop the spatial coordinates \( (x, y) \) to focus on the time evolution. The time derivative is denoted as \( \dot{u} \) (i.e., \( \frac{\partial u}{\partial t} \)), while spatial variation is captured through the gradient \( \nabla u \) (i.e., \( \frac{\partial u}{\partial x} \) and \( \frac{\partial u}{\partial y}  \)). Based on \cref{eq:ce}, we hypothesize that the atmospheric system follows the subsequent partial differential equation:

\vskip -0.2in

\begin{align} \label{eq:pde}
\dot{u}(t) &= \underbrace{- v(t) \cdot \nabla u(t) - u(t) \nabla \cdot v(t)}_{\text{Advection}} + s(t). 
\end{align}

\vskip -0.05in

Using the Method of Lines, we can express \cref{eq:pde} as a continuous first-order ODE system~\citep{climode}. In practice, the system is discretized into \( N \) time steps \( \{t_1, \ldots, t_N\} \), which allows us to leverage data from multiple future points to supervise the ODE in intermediate steps and apply numerical solvers like the Euler method~\citep{euler}. This results in the following discretized form:

\vskip -0.12in

\begin{align} \label{eq:solution}
    \begin{bmatrix}
     u(t_{n+1}) \\
     v(t_{n+1})
    \end{bmatrix}
    &= 
    \underbrace{
    \begin{bmatrix}
     u(t_n) \\
     v(t_n)
    \end{bmatrix}
    }_{\text{Initial Velocity $v(t_0)$}}
    + 
    \underbrace{
    \Delta t
    \begin{bmatrix}
    - \nabla \cdot (u(t_n) v(t_n)) \\
      \dot{v}(t_n)  \end{bmatrix}
     }_{\text{Advection ODE}}
     +
     \underbrace{
    \begin{bmatrix}
    s(t_n) \\
     0 \end{bmatrix}
     }_{\text{Source Term}}.
\end{align}
\vskip -0.1in
To solve this ODE system, three unknowns need to be estimated: \( v(t_0) \), \( \dot{v}(t_n) \), and \( s(t_n) \). As shown in \cref{fig:pipeline}, the proposed \NAME uses neural networks to model \( v(t_0) \) and \( s(t_n) \), and a neural ODE to model \( \dot{v}(t_n) \), which will be discussed in the following sections.

\subsection{Velocity Model}
\label{sec:velocity_model}

Modeling the initial velocity \( v(t_0) \) is crucial for ensuring the stability of the ODE solution. ClimODE~\citep{climode} estimates the initial velocity by first calculating the discrete-time derivative \( \frac{\Delta u}{\Delta t} \) from several past time points. However, using the discrete approximation \( \frac{\Delta u}{\Delta t} \) introduces large numerical errors, especially when \( \Delta t \) is not small enough. This approach struggles to capture smooth variations, resulting in significant deviations from the true continuous derivatives. Moreover, it involves a two-stage process where a separate model must first be trained to estimate all initial values \( v(t_0) \) before proceeding with the ODE solution.

Therefore, based on the following assumptions, we introduce the wave equation to leverage more precise spatial information for estimating the initial velocity.

\textbf{Incompressibility}: In this study, we assume that the fluid (air) behaves as incompressible. This implies that variations in pressure do not significantly influence the density of the fluid. This assumption is generally valid for large-scale weather phenomena; however, it may not be applicable to smaller, localized events.

\textbf{Linearization}: The governing equations of atmospheric dynamics can be linearized around a mean state, permitting the examination of small perturbations. This approach simplifies the mathematical framework and facilitates the superposition of solutions.

Given these assumptions, we can utilize the wave equation~\citep{pde}, commonly employed in atmospheric simulations, to enhance the estimation of the initial velocity based on the available spatial information, as outlined below:
% In contrast, \NAME leverages the wave equation, commonly used in atmospheric simulations, to improve the estimation of the initial velocity:
\begin{equation}
    \frac{\partial^2 u}{\partial t^2} = c^2 \left( \frac{\partial^2 u}{\partial x^2} + \frac{\partial^2 u}{\partial y^2} \right).
\end{equation}

This allows the first derivative with respect to time to be expressed as:
\begin{equation}
\frac{\partial u}{\partial t} = \int c^2 \left( \frac{\partial^2 u}{\partial x^2} + \frac{\partial^2 u}{\partial y^2} \right) dt. 
\end{equation}

Thus, \( \frac{\partial u}{\partial t} \) can be accurately computed as a function of the spatial derivatives \( \frac{\partial u}{\partial x} \) and \( \frac{\partial u}{\partial y} \), avoiding additional numerical errors. We model \( v(t_0) \) using a CNN-based neural network $f_v(\cdot)$:
\[
v(t_0) = f_v(u(t_0), \nabla u(t_0)).
\]

However, there is no free lunch, as we must also consider the discretization errors we introduce in the spatial domains. Coarse estimations based on 5.625° ERA5 data \citep{weatherbench} suggest a temporal resolution of $1/24$ and a spatial resolution of $1/(32*64)$, indicating that the spatial domain is nearly 100 times denser than the temporal domain. This disparity allows our approach to deliver a more precise and stable estimation of the initial velocity, which is vital for accurately solving the ODE system.

\subsection{Advection ODE}
\label{sec:advection_ode}

In the discretized ODE system in \cref{eq:solution}, the term \( \dot{u}(t_n) \) can be computed from the current values of \( u(t_n) \) and \( v(t_n) \) using the advection equation. For \( \dot{v}(t_n) \), we design an advection model:
\[
\dot{v}(t_n) = f_\theta(u(t_n), \nabla u(t_n), v(t_n), (\phi_s, \phi_t)),
\]
where \( (\phi_s, \phi_t) \) represent the spatial-temporal embeddings and details can be found in \cref{app:embeddings}.

The design of the advection model \( f_\theta \) is crucial for ensuring the stability of the numerical solution, as it takes the output from the velocity model as input. We argue that \( f_\theta \) should converge more slowly than the CNN-based velocity model, because the initial estimates of \( v(t_0) \) from the velocity model are likely to be inaccurate. If \( f_\theta \) converges too quickly based on early, imprecise values, it could cause the numerical solution to become unstable, potentially leading to failure during optimization. 

To address this, \( f_\theta \) is designed with a Vision Transformer (ViT)~\citep{vit} as the primary network, complemented by a linear term. The ViT, with its inherently slower convergence relative to CNNs, provides strong global modeling capabilities, while the linear term contributes to stabilizing the training process by promoting smoother convergence~\citep{stabilized}. A detailed analysis of how different architectural choices impact training stability is available in \cref{sec:abla_optimize}.

\subsection{Source Model}  
\label{sec:source_model}

To capture the energy gains and losses within the ODE system, we introduce a neural network to model the source term. Rather than incorporating the source term directly within the Advection ODE, we model it separately using the output of the Advection ODE \( \{u(t_n)\}_{n=1}^N \) to predict the corresponding source terms \( \{s(t_n)\}_{n=1}^N \). This approach mitigates the numerical errors that would arise from modeling \( s(t_i) \) within the ODE solver, as these errors would propagate through the solution. The source model \( f_s(\cdot) \) is formulated as follows:
\[
\{s(t_n)\}_{n=1}^N = f_s(\{u(t_n)\}_{n=1}^N, u(t_0), v(t_0), \phi_s, \{t_n\}_{n=1}^N).
\]

This model is supervised using the predicted values \( \{u(t_n)\}_{n=1}^N \), the spatial embedding \( \phi_s \), the sequence of time points \( \{t_n\}_{n=1}^N \), and the initial conditions \( u(t_0) \) and \( v(t_0) \). Rather than assuming the source term is independent at each time step, the model captures its temporal evolution, considering dependencies on both past and future values. The architecture of the source model is based on a 3D CNN, with further architectural details discussed in \cref{sec:abla_source}.

\subsection{SandWich Structure Design for solving advection equation}
\label{sec:sandwich}
The hybrid CNN-ViT-CNN architecture optimally combines local feature extraction and global context modeling, enabling efficient learning dynamics suited for distinct yet interconnected tasks in the advection equation estimation.

The sandwich design of our neural ODE model, comprising a CNN for fast-converging tasks (velocity estimation and source term modeling) and a ViT for slower-converging tasks (advection equation modeling), leverages the strengths of different architectures tailored to specific learning tasks. CNNs excel at local feature extraction and are particularly suited for tasks requiring rapid convergence, such as deriving initial conditions and identifying impacts from source terms with high spatial correlation. In contrast, Vision Transformers (ViTs) utilize attention mechanisms that capture global context and relationships, making them better suited for tasks with more complex interactions, such as solving the advection equation, where the dynamics often involve long-range dependencies. From a theoretical standpoint, the effectiveness of this hybrid architecture can be framed through the lens of inductive biases: the CNN's ability to model locality and translation invariance complements the ViT's ability to model global interactions and dependencies, resulting in a more robust solution strategy for the coupled problem. Moreover, such sandwich design choice is also related to the robustness of training as we discuss in \cref{sec:abla_optimize}.
% \subsection{Source Model}

% To account for energy gains and losses within the ODE system, we introduce a source network \( f_s \) to model the source term:

% \[
% u_{t_1:t_N} = f_s(u_{t_1:t_N}, u(t_0), v(t_0), \phi_s, \{t_1, \dots, t_N\})
% \]

% This model is supervised using the ODE-predicted values \( u_{t_1:t_N} \), the spatial embedding \( \psi(\x) \), the set of time points \( \{t_1, \dots, t_N\} \), and the initial conditions \( u_0 \) and \( v_0 \). Rather than treating the source term as independent across different time steps, the model captures its temporal evolution, considering the dependency of the source on both past and future values. This approach ensures a more accurate representation of the dynamic processes that drive energy fluctuations within the system.

\subsection{Multi-task Learning}
\label{sec:multi_task_learning}

Previous methods often train models using only the target leading time \( u(t_N) \) as the supervision signal, ignoring the valuable information contained in intermediate states \( \{u(t_n)\}_{n=1}^{N-1} \). Here, we adopt a multi-task learning strategy and leverage the continuous nature of neural ODE to predict the state at every intermediate time step \( \{u(t_n)\}_{n=1}^{N} \), minimizing the latitude-weighted RMSE between the predicted values \( u(t_n) \) and the ground truth \( \tilde{u}(t_n) \). The loss function is defined as:
\begin{equation}
\mathcal{L}=\frac{1}{N \times K \times H \times W} \sum_{n=1}^N \sum_{k=1}^K \sum_{h=1}^H \sum_{w=1}^W \alpha(h)\left(\tilde{u}_{k, h, w}(t_n) - u_{k, h, w}(t_n) \right)^2,
\end{equation}
where \( \alpha(h) \) is the latitude weighting factor that accounts for the varying grid cell areas on a spherical Earth, as cells near the equator cover larger areas than those near the poles. For more details on the weighting factor, refer to \cref{app:metric}.

By leveraging the multi-task learning strategy, the ODE system can exploit information across different time points, helping the model filter out errors arising from advection assumptions and neural network predictions. This allows us to train a single model with a lead time of \( N \) that can be used for inference at any time step up to \( N \), enhancing both efficiency and generalization.

% \begin{equation}
% \alpha(h)=\frac{\cos (\operatorname{lat}(h))}{\frac{1}{H} \sum_{h^{\prime}=1}^H \cos \left(\operatorname{lat}\left(h^{\prime}\right)\right)},
% \label{eq:wight}
% \end{equation}

% Here, \( \operatorname{lat}(h) \) denotes the latitude of the \( h \)-th row of the grid. The latitude weighting factor 

% compensates for the non-uniformity in grid cell areas on a spherical Earth, where cells near the equator cover larger areas than those near the poles, thus requiring higher weighting.

\section{Experiments}
\label{sec:exp}

In this section, we evaluate the proposed \NAME by forecasting the weather at a future time \( u(t+\Delta t) \) based on the conditions at a given time \( t \), where \( \Delta t \) (measured in hours) represents the lead time. The experimental setups are detailed in \cref{sec:setup}, while the results for global and regional weather forecasting are presented in \cref{sec:global_forecast} and \cref{sec:regional_forecast}, respectively.

\subsection{Experimental Setups}
\label{sec:setup}

\paragraph{Dataset.} We utilize the preprocessed ERA5 dataset from WeatherBench \citep{weatherbench}, which has 5.625° resolution (32 $\times$ 64 grid points) and temporal resolution of 1 hour. Our input data includes $K=48$ variables: 6 atmospheric variables at 7 pressure levels, 3 surface variables, and 3 constant fields. To evaluate the performance of \NAME, following the benchmark work in \cite{climode}, we focus on five target variables: geopotential at 500 hPa ($z500$), temperature at 850 hPa ($t850$), temperature at 2 meters ($t2m$), and zonal wind speeds at 10 meters ($u10$ and $v10$). We use the data from 1979 to 2015 as the training set, 2016 as the validation set, and 2017 to 2018 as the test set. 
% The $z500$ and $t850$ variables serve as standard benchmarks for medium-range NWP models and are widely used in previous deep learning studies, while $t2m$, $u10$, and $v10$ are crucial for applications relevant to human activities. We use data from 1979 to 2015 as the training set, 2016 as the validation set, and 2017 and 2018 as the test set. 
More details are available in \cref{app:era5}.

\paragraph{Metric.} In line with previous works, we evaluate all methods using latitude-weighted root mean squared error (RMSE) and latitude-weighted anomaly correlation coefficient (ACC):
\begin{align}
    \text{RMSE} &= \frac{1}{K} \sum_{k=1}^K \sqrt{\frac{1}{HW}\sum_{h=1}^H \sum_{w=1}^W \alpha(h)\left(\tilde{u}_{k, h, w}-u_{k, h, w}\right)^2}, \\
    \text{ACC} &= \frac{\sum_{k,h,w}\tilde{u}^{\prime}_{k, h, w} u^{\prime}_{k, h, w}}{\sqrt{\sum_{k,h,w} \alpha(h) (\tilde{u}^{\prime}_{k, h, w})^2 \sum_{k,h,w} \alpha(h) (u^{\prime}_{k, h, w})^2 }},
    \notag
\end{align}
where $\alpha(h)$ is the same latitude weighting factor as used in the training process; \( \tilde{u}^{\prime} = \tilde{u} - C \) and \( u^{\prime} = u - C \) are computed against the climatology \( C = \frac{1}{K}\sum_k \tilde{u}_k \), which is the temporal mean of the ground truth data over the entire test set. More details are available in \cref{app:metric}.

\paragraph{Baselines.} We compare \NAME with several representative methods from recent literature, including ClimaX~\citep{climax}, FourCastNet (FCN)~\citep{fourcastnet}, ClimODE~\citep{climode}, and the Integrated Forecasting System (IFS)~\citep{ifs}. Specifically, ClimaX is a pre-trained framework capable of learning from heterogeneous datasets that span different variables, spatial and temporal scales, and physical bases. FCN uses Adaptive Fourier Neural Operators to provide fast, high-resolution global weather forecasts. ClimODE is a physics-informed neural ODE model that incorporates key physical principles. IFS is the most advanced global physics simulation model of the European Center for Medium-Range Weather Forecasting (ECMWF).

\paragraph{Implementation details.} 
The architecture of our velocity model is based on ResNet2D \citep{resnet}, the ODE is based on ViT \citep{vit}, and the source model is based on ResNet3D. We optimize the model using the Adam optimizer.
% , with a learning rate of 5e-4 for both the velocity model and the source model, and 1e-4 for the advection model of the ODE system. 
Detailed discussions on the model architectures, specific parameter settings, and learning rate schedules are available in \cref{app:implement}.

\subsection{Global Weather Forecasting}
\label{sec:global_forecast}

\cref{tab:global_forecast} presents the global weather forecasting performance of \NAME and other baseline models at $\Delta t = \{6, 12, 18, 24\}$ hours. We report the results from the original ClimaX paper, where the model was pre-trained on the CMIP6 dataset~\citep{cmip6} and then fine-tuned on ERA5 dataset. Despite training solely on the ERA5 dataset, \NAME gains a 10\% improvement over ClimaX. 
Besides, \NAME surpasses ClimODE with a substantial improvement over 40\%, clearly demonstrating that we have effectively overcome the major challenges inherent in physics-driven weather forecasting models.
Furthermore, \NAME achieves performance on par with the IFS, which serves as the benchmark in the industry.

\subsection{Regional Weather Forecasting}
\label{sec:regional_forecast}

Global forecasting is not always feasible when only regional data is available. Therefore, we evaluate \NAME with other baselines for regional forecasting of relevant variables in North America, South America, and Australia, focusing on predicting future weather in each region based on its current conditions. The latitude boundaries for these regions are detailed in the \cref{app:regional}. As shown in \cref{tab:regional_forecast}, \NAME consistently achieves strong predictive performance across nearly all variables in each region, surpassing ClimaX and ClimODE by 59.7\% and 31.8\%, respectively. This underscores the strong ability of \NAME to model weather patterns effectively in data-scarce scenarios.

\footnotetext[1]{\label{climax_footnote}\url{https://github.com/microsoft/ClimaX}}

\renewcommand{\arraystretch}{1.15}
\setlength{\heavyrulewidth}{1.1pt}
\begin{table}[!t]
    \vskip -0.1in
    \centering
    \begin{threeparttable}
    \resizebox{\textwidth}{!}{
    \begin{tabular}{ccccccccc cccccc}
        \toprule
        & & \multicolumn{6}{c}{$\mathrm{RMSE} \ \downarrow$} & \multicolumn{6}{c}{$\mathrm{ACC} \ \uparrow$} \\
         \cmidrule(lr){3-8} \cmidrule(lr){9-14}
         
        \multirow{2}{*}{\scalebox{0.9}{Variable}} & \multirow{2}{*}{\scalebox{0.9}{Hours}} & \scalebox{0.7}{ClimaX $^\dagger$} & \scalebox{0.7}{FCN} & \scalebox{0.7}{IFS} & \scalebox{0.7}{ClimODE} & \scalebox{0.7}{\NAME} & \scalebox{0.7}{$\text{\NAME}^*$} & \scalebox{0.7}{ClimaX $^\dagger$} & \scalebox{0.7}{FCN} & \scalebox{0.7}{IFS} & \scalebox{0.7}{ClimODE} & \scalebox{0.7}{\NAME} & \scalebox{0.7}{$\text{\NAME}^*$} \\

         &  & \scalebox{0.7}{\citeyearpar{climax}} & \scalebox{0.7}{\citeyearpar{fourcastnet}} & \scalebox{0.7}{\citeyearpar{ifs}} & \scalebox{0.7}{\citeyearpar{climode}} & \scalebox{0.7}{(\textbf{Ours})} & \scalebox{0.7}{(\textbf{Ours})}
        & \scalebox{0.7}{\citeyearpar{climax}} & \scalebox{0.7}{\citeyearpar{fourcastnet}} & \scalebox{0.7}{\citeyearpar{ifs}} & \scalebox{0.7}{\citeyearpar{climode}} & \scalebox{0.7}{(\textbf{Ours})} & \scalebox{0.7}{(\textbf{Ours})} \\

        \toprule

        \multirow{4}{*}{$z500$} 
        
        & 6 & 62.7 & 149.4 & 26.9 & 102.9 & 54.0 & 56.3 & 1.00 & 0.99 & 1.00 & 0.99 & 1.00 & 1.00 \\
        & 12 & 81.9 & 217.8 & \textcolor{gray}{(N/A)} & 134.8 & 80.0 & 73.3 & 1.00 & 0.99 & \textcolor{gray}{(N/A)} & 0.99 & 1.00 & 1.00 \\
        & 18 & 88.9 & 275.0 & \textcolor{gray}{(N/A)} & 162.7 & 96.3 & 91.9 & 1.00 & 0.99 & \textcolor{gray}{(N/A)} & 0.98 & 1.00 & 1.00 \\
        & 24 & 96.2 & 333.0 & 51.0 & 193.4 & 114.5 & 114.5 & 1.00 & 0.99 & 1.00 & 0.98 & 1.00 & 1.00 \\

        \midrule

       \multirow{4}{*}{$t850$} 
        
        & 6 & 0.88 & 1.18 & 0.69 & 1.16 & 0.73 & 0.76 & 0.98 & 0.99 & 0.99 & 0.97 & 0.99 & 0.99 \\
        & 12 & 1.09 & 1.47 & \textcolor{gray}{(N/A)} & 1.32 & 0.87 & 0.88 & 0.98 & 0.99 & \textcolor{gray}{(N/A)} & 0.96 & 0.98 & 0.98 \\
        & 18 & 1.10 & 1.65 & \textcolor{gray}{(N/A)} & 1.47 & 0.95 & 0.95 & 0.98 & 0.99 & \textcolor{gray}{(N/A)} & 0.96 & 0.98 & 0.98 \\
        & 24 & 1.11 & 1.83 & 0.87 & 1.55 & 1.04 & 1.04 & 0.98 & 0.99 & 0.99 & 0.95 & 0.98 & 0.98 \\

        \midrule

        \multirow{4}{*}{$t2m$} 
        
        & 6 & 0.95 & 1.28 & 0.97 & 1.21 & 0.74 & 0.78 & 0.98 & 0.99 & 0.99 & 0.97 & 0.99 & 0.99 \\
        & 12 & 1.24 & 1.48 &\textcolor{gray}{(N/A)} & 1.45 & 0.88 & 0.89 & 0.97 & 0.99 & \textcolor{gray}{(N/A)} & 0.96 & 0.99 & 0.98 \\
        & 18 & 1.19 & 1.61 & \textcolor{gray}{(N/A)} & 1.43 & 0.95 & 0.95 & 0.97 & 0.99 & \textcolor{gray}{(N/A)} & 0.96 & 0.98 & 0.98 \\
        & 24 & 1.10 & 1.68 & 1.02 & 1.40 & 0.98 & 0.98 & 0.98 & 0.99 & 0.99 & 0.96 & 0.98 & 0.98 \\

        \midrule

        \multirow{4}{*}{$u10$} 
        
        & 6 & 1.08 & 1.47 & 0.80 & 1.41 & 0.84 & 0.88 & 0.97 & 0.95 & 0.98 & 0.91 & 0.98 & 0.98 \\
        & 12 & 1.23 & 1.89 & \textcolor{gray}{(N/A)} & 1.81 & 1.00 & 1.00 & 0.95 & 0.93 & \textcolor{gray}{(N/A)} & 0.89 & 0.97 & 0.97 \\
        & 18 & 1.27 & 2.05 &\textcolor{gray}{(N/A)} & 1.97 & 1.12 & 1.13 & 0.95 & 0.91 & \textcolor{gray}{(N/A)} & 0.88 & 0.96 & 0.96 \\
        & 24 & 1.41 & 2.33 & 1.11  & 2.01 & 1.26 & 1.26 & 0.94 & 0.89 & 0.97 & 0.87 & 0.95 & 0.95 \\

        \midrule

        \multirow{4}{*}{$v10$} 
        
        & 6 & \textcolor{gray}{(N/A)} & 1.54 & 0.94 & 1.53 & 0.87 & 0.90 & \textcolor{gray}{(N/A)} & 0.94 & 0.98 & 0.92 & 0.98 & 0.98 \\
        & 12 & \textcolor{gray}{(N/A)} & 1.81 & \textcolor{gray}{(N/A)} & 1.81 & 1.04 & 1.04 & \textcolor{gray}{(N/A)} & 0.91 & \textcolor{gray}{(N/A)} & 0.89 & 0.97 & 0.97 \\
        & 18 & \textcolor{gray}{(N/A)} & 2.11 &\textcolor{gray}{(N/A)} & 1.96 & 1.15 & 1.16 & \textcolor{gray}{(N/A)} & 0.86 & \textcolor{gray}{(N/A)} & 0.88 & 0.96 & 0.96 \\
        & 24 & \textcolor{gray}{(N/A)} & 2.39 & 1.33 & 2.04 & 1.29 & 1.29 & \textcolor{gray}{(N/A)} & 0.83 & 0.97 & 0.86 & 0.95 & 0.95 \\

        \bottomrule
    \end{tabular}}

    \begin{tablenotes}
        \tiny
        \item [$\dagger$] For 6h and 24h, we report results from the original ClimaX paper; 12h and 18h results are obtained using their official pre-trained model and code\hyperref[climax_footnote]{\textsuperscript{1}}.
        \item [*] Indicates a 24-hour model used for inference across all lead times.
    \end{tablenotes}
    \caption{\small Latitude-weighted RMSE and ACC comparison with baseline models for various target variables across different lead times on global weather forecasting.}
    \label{tab:global_forecast}
    \end{threeparttable}
    \vskip -0.1in
\end{table}

\begin{table}[!t]
    \centering
    \begin{threeparttable}
    \resizebox{\textwidth}{!}{
    \begin{tabular}{lccccccc cccc cccc}
        \toprule
        & & \multicolumn{4}{c}{North-America} &  \multicolumn{4}{c}{South-America} &  \multicolumn{4}{c}{Australia}   \\
        \cmidrule(lr){3-6} \cmidrule(lr){7-10} \cmidrule(lr){11-14} 
        
        \multirow{2}{*}{\scalebox{0.9}{Variable}} & \multirow{2}{*}{\scalebox{0.9}{Hours}} & \scalebox{0.7}{ClimaX$^\dagger$} & \scalebox{0.7}{ClimODE} & \scalebox{0.7}{\NAME} & \scalebox{0.7}{$\text{\NAME}^*$} & \scalebox{0.7}{ClimaX$^\dagger$} & \scalebox{0.7}{ClimODE} & \scalebox{0.7}{\NAME} & \scalebox{0.7}{$\text{\NAME}^*$} & \scalebox{0.7}{ClimaX$^\dagger$} & \scalebox{0.7}{ClimODE} & \scalebox{0.7}{\NAME} & \scalebox{0.7}{$\text{\NAME}^*$} \\

        &  & 
        \scalebox{0.7}{\citeyearpar{climax}} & \scalebox{0.7}{\citeyearpar{climode}} & \scalebox{0.7}{(\textbf{Ours})} & \scalebox{0.7}{(\textbf{Ours})} & 
        \scalebox{0.7}{\citeyearpar{climax}} & \scalebox{0.7}{\citeyearpar{climode}} & \scalebox{0.7}{(\textbf{Ours})} & \scalebox{0.7}{(\textbf{Ours})} & 
        \scalebox{0.7}{\citeyearpar{climax}} & \scalebox{0.7}{\citeyearpar{climode}} & \scalebox{0.7}{(\textbf{Ours})} & \scalebox{0.7}{(\textbf{Ours})} \\

        \toprule

        \multirow{4}{*}{$z500$} 
        & 6  & 273.4 & 134.5 & 91.2  & 97.3  & 205.4 & 107.7 & 62.3  & 68.9  & 190.2 & 103.8 & 62.7  & 58.4 \\
        & 12 & 329.5 & 225.0 & 147.4 & 158.7 & 220.2 & 169.4 & 97.7  & 100.0 & 184.7 & 170.7 & 79.2  & 77.7 \\
        & 18 & 543.0 & 307.7 & 218.9 & 233.5 & 269.2 & 237.8 & 137.5 & 141.2 & 222.2 & 211.1 & 103.5 & 102.7 \\
        & 24 & 494.8 & 390.1 & 314.5 & 314.5 & 301.8 & 292.0 & 183.1 & 183.1 & 324.9 & 308.2 & 125.1 & 125.1 \\

        \midrule

       \multirow{4}{*}{$t850$} 
        & 6  & 1.62 & 1.28 & 0.88 & 0.94 & 1.38 & 0.97 & 0.73 & 0.77 & 1.19 & 1.05 & 0.65 & 0.64 \\
        & 12 & 1.86 & 1.81 & 1.09 & 1.15 & 1.62 & 1.25 & 0.91 & 0.92 & 1.30 & 1.20 & 0.76 & 0.76 \\
        & 18 & 2.75 & 2.03 & 1.28 & 1.35 & 1.79 & 1.43 & 1.06 & 1.07 & 1.39 & 1.33 & 0.87 & 0.86 \\
        & 24 & 2.27 & 2.23 & 1.57 & 1.57 & 1.97 & 1.65 & 1.25 & 1.25 & 1.92 & 1.63 & 0.97 & 0.97 \\

        \midrule

        \multirow{4}{*}{$t2m$} 
        & 6  & 1.75 & 1.61 & 0.66 & 0.71 & 1.85 & 1.33 & 0.80 & 0.86 & 1.57 & 0.80 & 0.73 & 0.71 \\
        & 12 & 1.87 & 2.13 & 0.78 & 0.84 & 2.08 & 1.04 & 0.96 & 0.98 & 1.57 & 1.10 & 0.81 & 0.81 \\
        & 18 & 2.27 & 1.96 & 0.86 & 0.93 & 2.15 & 0.98 & 1.07 & 1.08 & 1.72 & 1.23 & 0.89 & 0.88 \\
        & 24 & 1.93 & 2.15 & 0.99 & 0.99 & 2.23 & 1.17 & 1.17 & 1.17 & 2.15 & 1.25 & 0.93 & 0.93 \\

        \midrule

        \multirow{4}{*}{$u10$} 
        & 6  & 1.74 & 1.54 & 1.05 & 1.09 & 1.27 & 1.25 & 0.83 & 0.87 & 1.40 & 1.35 & 1.02 & 1.04 \\
        & 12 & 2.24 & 2.01 & 1.37 & 1.42 & 1.57 & 1.49 & 1.05 & 1.03 & 1.77 & 1.78 & 1.24 & 1.27 \\
        & 18 & 3.24 & 2.17 & 1.77 & 1.81 & 1.83 & 1.81 & 1.19 & 1.20 & 2.03 & 1.96 & 1.39 & 1.45 \\
        & 24 & 3.14 & 2.34 & 2.22 & 2.22 & 2.04 & 2.08 & 1.39 & 1.39 & 2.64 & 2.33 & 1.62 & 1.62 \\

        \midrule

        \multirow{4}{*}{$v10$} 
        & 6  & 1.83 & 1.67 & 1.12 & 1.16 & 1.31 & 1.30 & 0.89 & 0.92 & 1.47 & 1.44 & 1.09 & 1.10 \\
        & 12 & 2.43 & 2.03 & 1.52 & 1.57 & 1.64 & 1.71 & 1.11 & 1.10 & 1.79 & 1.87 & 1.28 & 1.32 \\
        & 18 & 3.52 & 2.31 & 2.00 & 2.05 & 1.90 & 2.07 & 1.26 & 1.28 & 2.33 & 2.23 & 1.41 & 1.48 \\
        & 24 & 3.39 & 2.50 & 2.56 & 2.56 & 2.14 & 2.43 & 1.49 & 1.49 & 2.58 & 2.53 & 1.64 & 1.64 \\
        \bottomrule
    \end{tabular}}
    \begin{tablenotes}
        \tiny
        \item [$\dagger$] The number is cited from ClimODE \citep{climode}.
    \end{tablenotes}
    \caption{\small Latitude-weighted RMSE comparison with baseline models for various target variables across different lead times on regional weather forecasting.}
    \vskip -0.1in
    \label{tab:regional_forecast}
    \end{threeparttable}
\end{table}

\subsection{Flexible Inference with a Single 24-Hour Model}

Many deep learning-based methods treat predictions for different lead times as separate tasks, requiring a distinct model for each lead time. Some approaches attempt to use short-range models with rolling strategies~\citep{pangu, fengwu}, but they still face the challenge of error accumulation. In contrast, by modeling the atmosphere as a physics-driven continuous process and designing a time-dependent source network to account for errors at each time step, \NAME can capture information across all intermediate time points. As shown in \cref{tab:global_forecast} and \cref{tab:regional_forecast}, $\text{\NAME}^*$ (a 24-hour model of \NAME used for inference across all lead times) demonstrates its effectiveness for any hour within that period. The results show that $\text{\NAME}^*$ achieves performance comparable to \NAME across most variables and even exceeds \NAME in certain cases (e.g., $z500$). This highlights the effectiveness of our physics-driven ODE model in filtering out accumulated errors.

\section{Ablation Studies}

\subsection{Effectiveness of Wave Equation-Informed Estimation}

\begin{figure}[!t]
    \centering
    \includegraphics[width=\textwidth]{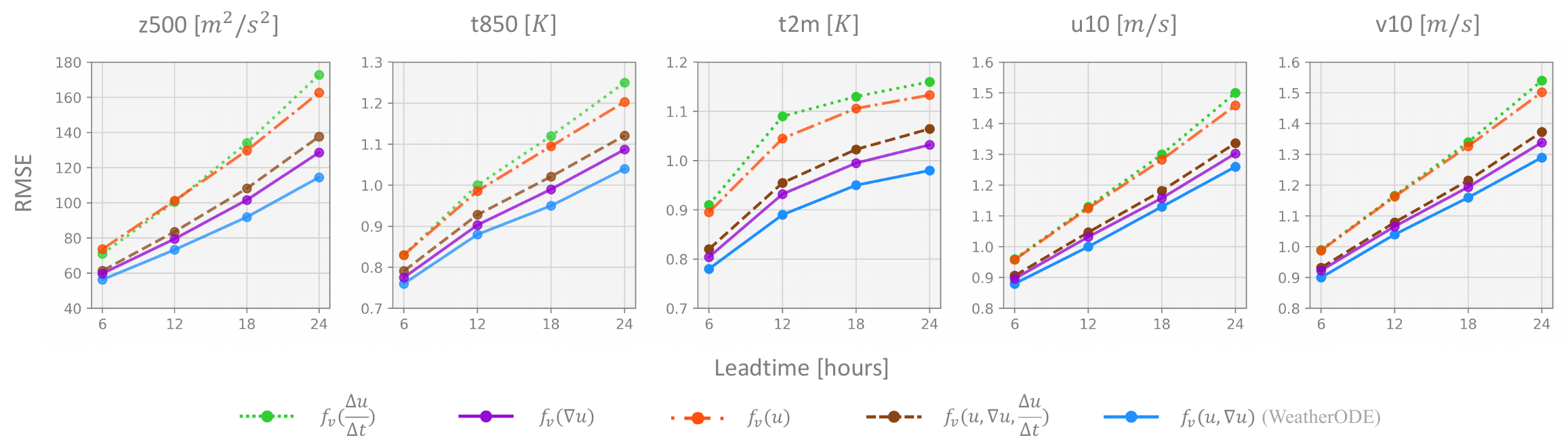}
    \caption{RMSE comparison for different input configurations of the velocity model.}
    \label{fig:ablation_v_net}
\end{figure}

\begin{figure}[!t]
    \centering
    \includegraphics[width=1\linewidth]{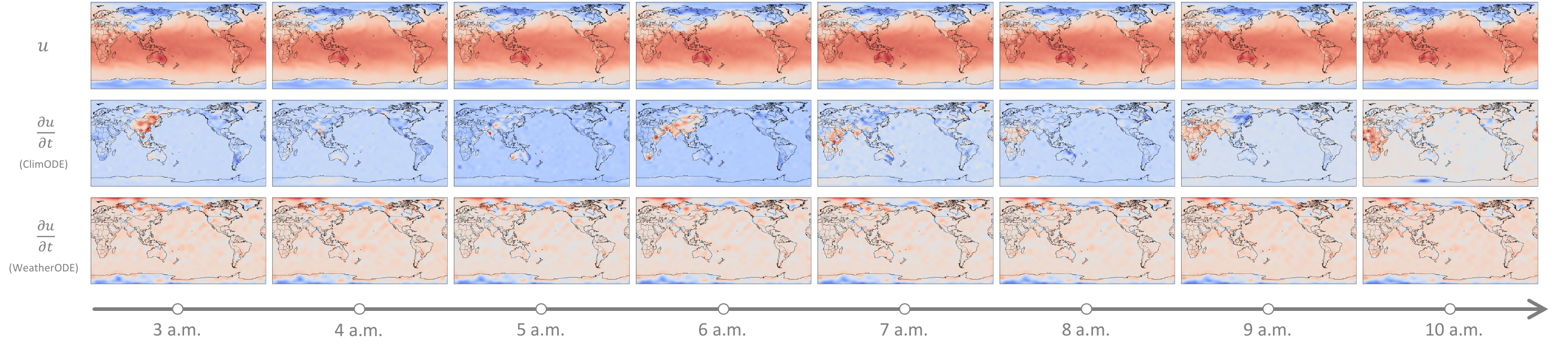}
    \caption{\small Visualization of the 2-meter temperature \( u \) on January 1, 2017, from 3 a.m. to 10 a.m., with the estimated \( \frac{\partial u}{\partial t} \) from ClimODE and \NAME. \NAME provides smoother, more continuous estimates of \( \frac{\partial u}{\partial t} \), closely matching \( u \), while ClimODE shows abrupt changes.}
    \vskip -0.2in
    \label{fig:vis_partial_u_t}
\end{figure}

To validate the superiority of the wave equation-informed estimation over the discrete-time derivative, we conduct five experiments of the velocity model to estimate the initial velocity: 
(1) $f_v(\frac{\Delta u}{\Delta t})$: the model uses only the discrete-time derivative $\frac{\Delta u}{\Delta t}$; 
(2) $f_v(u, \nabla u, \frac{\Delta u}{\Delta t})$: the model combines the discrete-time derivative with $u$ and $\nabla u$;
(3) $f_v(u)$: the model uses only $u$;
(4) $f_v(\nabla u)$: the model uses only $\nabla u$;
(5) $f_v(u, \nabla u)$: the model relies solely on the wave function-derived $u$ and $\nabla u$. 
The results in \cref{fig:ablation_v_net} demonstrate the effectiveness of the wave equation-informed approach. Specifically, (1) has an RMSE that is over 20\% worse compared to (5). It is notable that experiment the incorporation of $\frac{\Delta u}{\Delta t}$ into the velocity model in (2) adversely affected performance compared to (5),  primarily due to overfitting arising from the substantial discrepancy between the discrete-time derivative and the true values. Furthermore, the model in (5) outperforms (4), suggesting that the inclusion of $\nabla u$ with $u$ provides additional beneficial information to the network, enhancing its predictive capability. \cref{fig:vis_partial_u_t}  shows that \NAME produces much smoother $\frac{\Delta u}{\Delta t}$ predictions, aligning with the smooth nature of $u$, while the predictions of ClimODE are more erratic.

\subsection{Analysis of Source Model Architecture}
\label{sec:abla_source}

\begin{figure}[!t]
    \centering
    \includegraphics[width=\textwidth]{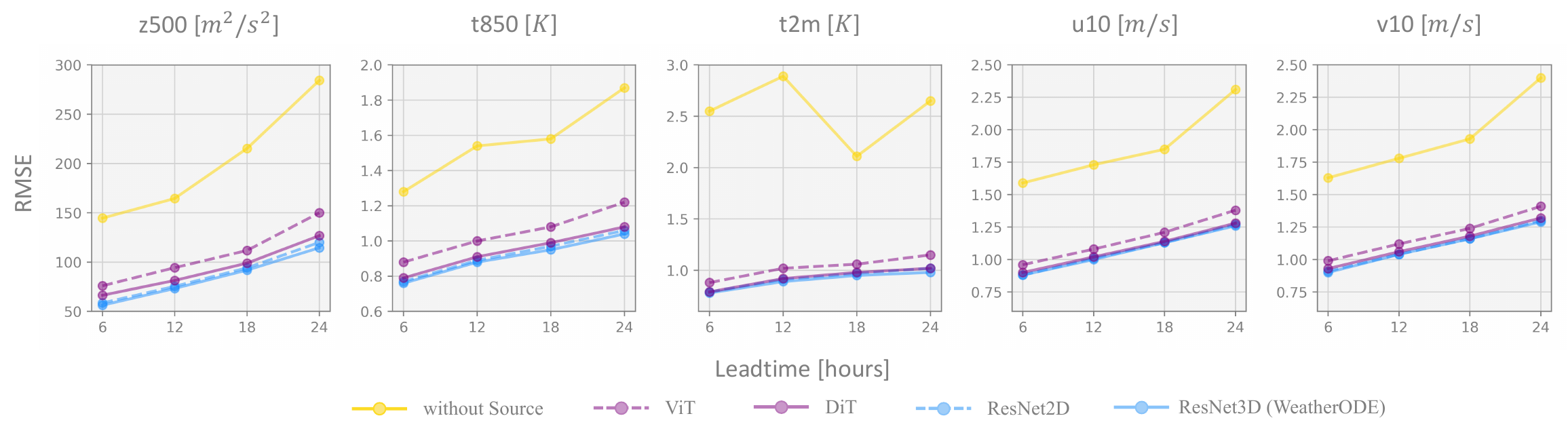}
    \caption{RMSE comparison for different architectures of the source model.}
    \label{fig:ablation_source_net}
    \vskip -0.2in
\end{figure}
We conduct experiments by removing the source model and comparing different source model architectures: ViT, DiT, ResNet2D, and ResNet3D. DiT~\citep{dit} and ResNet3D are the time-aware versions of ViT and ResNet2D, respectively.
 As shown in \cref{fig:ablation_source_net}, DiT and ResNet3D outperform ViT and ResNet2D by 10\% and 5\%, and significantly exceed the performance of the model without the source component. These results demonstrate the effectiveness of the source model and highlight the importance of integrating temporal information into its architecture.

\subsection{Stability Analysis of Neural Network and Neural ODE Integration}
\label{sec:abla_optimize}

The interdependencies between the advection and velocity models highlight the importance of carefully selecting architectures and learning rates to ensure the stability and performance of the neural network and neural ODE system. As shown in \cref{tab:app_stability}, the learning rate for the advection model must be lower than that of the velocity model due to often inaccurate initial estimates. If the advection model converges too quickly based on these estimates, it may lead to numerical instabilities and NaN values. Alternatively, using an advection model architecture with inherently slower convergence can yield similar results even with the same learning rate. Moreover, given that the source term represents solar energy with strong locality—where energy patterns are similar in neighboring regions—a CNN architecture that effectively captures local dependencies is ideal for this task.

\renewcommand{\arraystretch}{1.2}
\begin{table}[htb]
    \centering
    \resizebox{\textwidth}{!}{
    \begin{tabular}{ccccccc}
    \toprule
     \textbf{Velocity Model} & \textbf{Advection Model} & \textbf{Source Model} & \textbf{lr}    & \textbf{Advection lr} & \textbf{Training Stable?} & \textbf{Rank} \\
    \toprule 

    CNN            & ViT       & CNN          &  5e-4  & 5e-4   & \textcolor{green}{\correct} & 1  \\ 
    ViT            & ViT       & CNN          &  5e-4  & 5e-4   & \textcolor{green}{\correct} & 4 \\ 
    CNN            & ViT       & ViT          &  5e-4  & 5e-4   & \textcolor{green}{\correct} & 2 \\ 
    ViT            & ViT       & ViT          &  5e-4  & 5e-4   & \textcolor{green}{\correct} & 5  \\ 
    CNN            & CNN       & CNN          &  5e-4  & 5e-4   & \textcolor{red}{\wrong} (1) & - \\ 
    ViT            & CNN       & CNN          &  5e-4  & 5e-4   & \textcolor{red}{\wrong} (1) & - \\ 
    CNN            & CNN       & ViT          &  5e-4  & 5e-4   & \textcolor{red}{\wrong} (1) & - \\ 
    ViT            & CNN       & ViT          &  5e-4  & 5e-4   & \textcolor{red}{\wrong} (1) & - \\ 
    CNN            & CNN       & CNN          &  5e-4  & 5e-5   & \textcolor{green}{\correct} & 3 \\ 
    ViT            & CNN       & ViT          &  5e-4  & 5e-5   & \textcolor{red}{\wrong} (3) & - \\ 
    ViT            & CNN       & ViT          &  5e-4  & 5e-6   & \textcolor{green}{\correct} & 6 \\ 
    
    \toprule

    \end{tabular}
    }
    \caption{\small Stability analysis of neural network and neural ODE integration across different architectures and learning rates. ``Advection lr'' denotes the learning rate of the advection model and ``lr'' corresponds to the other two. ``\textcolor{green}{\correct}'' indicates stable training, and ``\textcolor{red}{\wrong} ($i$)'' shows where NaN values occurred at epoch $i$. ``Rank'' indicates the performance ranking among stable configurations.}
    \vskip -0.1in
    \label{tab:app_stability}
\end{table}

% See \cref{app:stability} for the full result.

% \begin{table}[htb]
%     \centering
%     \footnotesize
%     \resizebox{\textwidth}{!}{
%     \begin{tabular}{ccccccc}
%     \toprule
%      \textbf{Velocity Model} & \textbf{Advection Model} & \textbf{Source Model} & \textbf{lr}    & \textbf{Advection lr} & \textbf{Training Stable?} & \textbf{Rank} \\
%     \toprule 

%     CNN            & ViT       & CNN &  5e-4  & 5e-4   & \textcolor{green}{\correct} & 1  \\ 
%     CNN            & CNN       & CNN &  5e-4  & 5e-4   & \textcolor{red}{\wrong} (1) & - \\ 
%     CNN            & CNN       & CNN & 5e-4  & 5e-5   & \textcolor{green}{\correct} & 2 \\ 
%     ViT            & CNN       & ViT &  5e-4  & 5e-4   & \textcolor{red}{\wrong} (1) & - \\ 
%     ViT            & CNN       & ViT & 5e-4  & 5e-5   & \textcolor{red}{\wrong} (3) & - \\ 
%     ViT            & CNN       & ViT & 5e-4  & 5e-6   & \textcolor{green}{\correct} & 3 \\ 
    
%     \toprule

%     \end{tabular}
%     }
%     \caption{Stability analysis of neural network and neural ODE integration across different architectures and learning rates. ``Advection lr'' denotes the learning rate of the advection model and ``lr'' corresponds to the other two. ``\textcolor{green}{\correct}'' indicates stable training, and ``\textcolor{red}{\wrong} ($i$)'' shows where NaN values occurred at epoch $i$. ``Rank'' indicates the performance ranking among stable configurations.}
%     \label{tab:abala_optimize}
% \end{table}

\section{Conclusion}
In this paper, we tackle several challenges faced by neural ODE-based weather forecasting models, specifically addressing time-discretization errors, global-local biases across individual tasks in solving the advection equation, and discrepancies in time-dependent sources that compromise predictive accuracy. To address these issues, we present WeatherODE—a novel sandwich neural ODE model that integrates wave equation theory with a dynamic source model. This approach effectively reduces errors and promotes synergy between neural networks and neural ODEs. Our in-depth analysis of WeatherODE's architecture and optimization establishes a strong foundation for advancing hybrid modeling in meteorology. Looking forward, our work opens avenues for further exploration of hybrid models that blend traditional physics-driven and modern machine-learning techniques.

\clearpage
\bibliography{iclr2025_conference}
\bibliographystyle{iclr2025_conference}

\clearpage

\appendix

\section{ERA5 Data}
\label{app:era5}

We train \NAME on the preprocessed $5.625$° ERA5 data from WeatherBench \citep{weatherbench}, a benchmark dataset and evaluation framework designed to facilitate the comparison of data-driven weather forecasting models. WeatherBench regridded the raw ERA5 dataset\footnote{For more details of the raw ERA5 data, see \url{https://confluence.ecmwf.int/display/CKB/ERA5\%3A+data+documentation}.} from its $0.25$° resolution to three coarser resolutions: $5.625$°, $2.8125$°, and $1.40625$°. The processed dataset includes 8 atmospheric variables across 13 pressure levels, 6 surface variables, and 5 static variables. For training and testing \NAME, we selected 6 atmospheric variables at 7 pressure levels, 3 surface variables, and 3 static variables, as detailed in \cref{tab:met_variables_summary}. 
% We focus on five target variables: geopotential at 500 hPa ($z500$), temperature at 850 hPa ($t850$), temperature at 2 meters ($t2m$), and zonal wind speeds at 10 meters ($u10$ and $v10$). 

\begin{table}[htbp]
\centering
\renewcommand{\arraystretch}{1.3}
\resizebox{\textwidth}{!}{
\begin{tabular}{@{}llp{8cm}p{5.5cm}@{}}
\toprule
\textbf{Variable Name} & \textbf{Abbrev.} & \textbf{Description} & \textbf{Levels} \\ 

\toprule
 
Land-sea mask               & $lsm$       & Binary mask distinguishing land (1) from sea (0)      & N/A            \\
Orography                   & $oro$       & Height of Earth's surface                             & N/A            \\
Latitude                    & $lat$       & Latitude of each grid point                           & N/A            \\
% Longitude                   & $lon$       & Longitude of each grid point                          & N/A            \\

\midrule
\midrule

2 metre temperature         & $t2m$       & Temperature measured 2 meters above the surface       & Single level   \\
10 metre U wind component   & $u10$       & East-west wind speed at 10 meters above the surface   & Single level   \\
10 metre V wind component   & $v10$       & North-south wind speed at 10 meters above the surface & Single level   \\ 
Geopotential                & $z$         & Height relative to a pressure level                   & $50, 250, 500, 600, 700, 850, 925$ hPa \\
U wind component            & $u$         & Wind speed in the east-west direction                 & $50, 250, 500, 600, 700, 850, 925$ hPa \\
V wind component            & $v$         & Wind speed in the north-south direction               & $50, 250, 500, 600, 700, 850, 925$ hPa \\
Temperature                 & $t$         & Atmospheric temperature                               & $50, 250, 500, 600, 700, 850, 925$ hPa \\
Specific humidity           & $q$         & Mixing ratio of water vapor to total air mass         & $50, 250, 500, 600, 700, 850, 925$ hPa \\
Relative humidity           & $r$         & Humidity relative to saturation                       & $50, 250, 500, 600, 700, 850, 925$ hPa \\ \bottomrule
\end{tabular}
}

\caption{Summary of ECMWF variables utilized in the ERA5 dataset. The variables $lsm$ and $oro$ are constant and invariant with time, while $t2m$, $u10$, and $v10$ represent surface variables. The remaining are atmospheric variables which are measured at specific pressure levels.}
\label{tab:met_variables_summary}
\end{table}

\section{Weather Forecasting Metrics}
\label{app:metric}

In this section, we provide a detailed explanation of all the evaluation metrics used in \cref{sec:exp}. For each metric, \( u \) and \( \tilde{u} \) represent the predicted and ground truth values, respectively, both shaped as \( K \times H \times W \), where \( K \) is the number of predict quantities, and \( H \times W \) is the spatial resolution. The latitude weighting term \( \alpha(\cdot) \) accounts for the non-uniform grid cell areas.

\paragraph{\textbf{Latitude-weighted Root Mean Square Error (RMSE)}} assesses model accuracy while considering the Earth's curvature. The latitude weighting adjusts for the varying grid cell areas at different latitudes, ensuring that errors are appropriately measured. Lower RMSE values indicate better model performance.

\[
\text{RMSE} = \frac{1}{K} \sum_{k=1}^K \sqrt{\frac{1}{HW}\sum_{h=1}^H \sum_{w=1}^W \alpha(h)\left(\tilde{u}_{k, h, w}-u_{k, h, w}\right)^2}, \ \alpha(h) = \frac{\cos (\operatorname{lat}(h))}{\frac{1}{H} \sum_{h^{\prime}=1}^H \cos \left(\operatorname{lat}\left(h^{\prime}\right)\right)}.
\]

% \[
% \alpha(h) = \frac{\cos (\operatorname{lat}(h))}{\frac{1}{H} \sum_{h^{\prime}=1}^H \cos \left(\operatorname{lat}\left(h^{\prime}\right)\right)}
% \]

\paragraph{\textbf{Anomaly Correlation Coefficient (ACC)}} measures a model's ability to predict deviations from the mean. Higher ACC values indicate better accuracy in capturing anomalies, which is crucial in meteorology and climate science.

\[
\text{ACC} = \frac{\sum_{k,h,w}\tilde{u}^{\prime}_{k, h, w} u^{\prime}_{k, h, w}}{\sqrt{\sum_{k,h,w} \alpha(h) (\tilde{u}^{\prime}_{k, h, w})^2 \sum_{k,h,w} \alpha(h) (u^{\prime}_{k, h, w})^2 }},
\]

where \( u^{\prime} = u - C \) and \( \tilde{u}^{\prime} = \tilde{u} - C \), with \( C = \frac{1}{K}\sum_k \tilde{u}_k \) representing the temporal mean of the ground truth over the test set.

\section{Implementation Details}
\label{app:implement}

\subsection{Data Flow}

We normalized all inputs by computing the mean and standard deviation for each variable at each pressure level (for atmospheric variables) to achieve zero mean and unit variance. After normalization, the input \( u(t_0) \in \mathbb{R}^{K \times H \times W} \) with its spatial derivative \( \nabla u(t_0) \in \mathbb{R}^{2K \times H \times W} \) are processed by the velocity model $f_v(\cdot)$ to estimate the initial velocity \( v_0 \in \mathbb{R}^{2K \times H \times W} \). Both \( u(t_0) \) and \( v(t_0) \) are then fed into the ODE system, where the \( \dot{u}(t_n) \) is calculated by advection equation and the advection model $f_\theta(\cdot)$ uses \(u(t_n), \nabla u(t_n), v(t_n) \) and \( (\phi_s, \phi_t)\) to model \( \dot{v}(t_n) \). The ODE system outputs the predicted future state \( \{ u(t_n) \}_{n=1}^{N} \), where \( N \) represents the lead time.

The predicted \( \{ u(t_n) \}_{n=1}^{N} \), along with \( u(t_0) \), and \( v(t_0) \), are then passed into the source model $f_s(\cdot)$ to estimate the source term  \( \{ s(t_n) \}_{n=1}^{N} \). The final prediction for the lead time and each intermediate time point is obtained by adding \( s(t_n) \) to \( u(t_n) \) and then applying the inverse normalization. For training and evaluation, we selected five key variables from the \( K \) input variables: $z500$, $t850$, $t2m$, $u10$, and $v10$. 
%The $z500$ and $t850$ variables serve as standard benchmarks for medium-range NWP models and are widely used in previous deep learning studies, while $t2m$, $u10$, and $v10$ are crucial for applications relevant to human activities. 

\subsection{Embeddings}
\label{app:embeddings}

\paragraph{Spatial Encoding} Latitude \( h \) and longitude \( w \) are encoded with trigonometric and spherical coordinates:

\vskip -0.1in

\[
\phi_s = \left[ \sin(h), \cos(h), \sin(w), \cos(w), \sin(h) \cos(w), \sin(h) \sin(w) \right].
\]

\paragraph{Temporal Encoding} Daily and seasonal cycles are encoded using trigonometric functions:

\[
\phi_t = \left[ \sin(2\pi t), \cos(2\pi t), \sin\left(\frac{2\pi t}{365}\right), \cos\left(\frac{2\pi t}{365}\right) \right].
\]

\paragraph{Spatiotemporal Embedding} The final embedding integrates both:

\vskip -0.1in

\[
(\phi_s, \phi_t) = \left[\phi_s, \phi_t, \phi_s \times \phi_t \right].
\]

\subsection{Optimization}

All experiments are conducted with a batch size of $8$, running on $4$ NVIDIA A800-SXM4-80GB GPUs for $50$ epochs. We use the AdamW optimizer with \( \beta_1 = 0.9 \), \( \beta_2 = 0.999 \). The learning rate is set to $1e$-$4$ for the ODE model components and $5e$-$4$ for the rest. A weight decay of $1e$-$5$ is applied to all parameters except for the positional embeddings. The learning rate follows a linear warmup schedule starting from $1e$-$8$ for the first $10,000$ steps (approximately $1$ epoch), transitioning to a cosine-annealing schedule for the remaining $90,000$ steps (approximately $9$ epochs), with a minimum value of $1e$-$8$.

\subsection{Hyperparameters}

\begin{table}[htb]
    \centering
    \small
    \setlength{\tabcolsep}{5pt}
    \begin{tabular}{lll}
    \toprule
    \textbf{Hyperparameter} & \textbf{Description} & \textbf{Value}     \\ 
    
    \toprule

    Kernel size            & Size of each convolutional kernels             & $3$   \\
    Padding size           & Size of padding of each convolutional layer    & $1$   \\
    Stride                 & Step size of each convolutional layer          & $1$   \\
    Dropout                & Dropout probability                            & $0.1$ \\
    Leakage Coefficient    & Slope of LeakyReLU for negative inputs         & $0.3$ \\
    ResBlock List          & (Number of ResBlocks, Hidden dimensions)       & $[(5, 512), (5, 128), (3, 64), (2, 48)]$   \\
    \toprule
    \end{tabular}
    \caption{Default hyperparameters of ResNet2D of velocity model.}
    \label{tab:velocity_model}
\end{table}

\begin{table}[htb]
    \centering
    \small
    \setlength{\tabcolsep}{8pt}
    \begin{tabular}{lll}
    \toprule
    \textbf{Hyperparameter} & \textbf{Description} & \textbf{Value}     \\ 
    
    \toprule

    $p$                    & Size of image patches         & $2$   \\
    $D$                    & Dimension of hidden layers    & $1024$   \\
    Depth                  & Number of Transformer blocks  & $4$   \\
    Heads                  & Number of attention heads     & $8$ \\
    MLP ratio              & Expansion factor for MLP      & $4$ \\
    Decoder Depth          & Number of layers of the final prediction head & $2$  \\
    Drop path              & Stochastic depth rate         & $0.1$ \\
    Dropout                & Dropout rate                  & $0.1$ \\
    \toprule
    \end{tabular}

    \caption{Default hyperparameters of ViT in advection ODE.}
    \label{tab:vit_model}
\end{table}

\begin{table}[htb]
    \centering
    \small
    \setlength{\tabcolsep}{5pt}
    \begin{tabular}{lll}
    \toprule
    \textbf{Hyperparameter} & \textbf{Description} & \textbf{Value}     \\ 
    
    \toprule

    Kernel size            & Size of each 3D convolutional kernels             & $3$   \\
    Padding size           & Size of padding of each 3D convolutional layer    & $1$   \\
    Stride                 & Step size of each 3D convolutional layer          & $1$   \\
    Dropout                & Dropout probability                            & $0.1$ \\
    Leakage Coefficient    & Slope of LeakyReLU for negative inputs         & $0.3$ \\
    ResBlock List          & (Number of ResBlocks, Hidden dimensions)       & $[(5, 512), (5, 128), (3, 64), (2, 48)]$   \\
    \toprule
    \end{tabular}
    \caption{Default hyperparameters of ResNet3D of source model.}
    \label{tab:source_model}
\end{table}

\section{Regional Forecast}
\label{app:regional}

\begin{figure}[htb]
    \centering
    \includegraphics[width=1\linewidth]{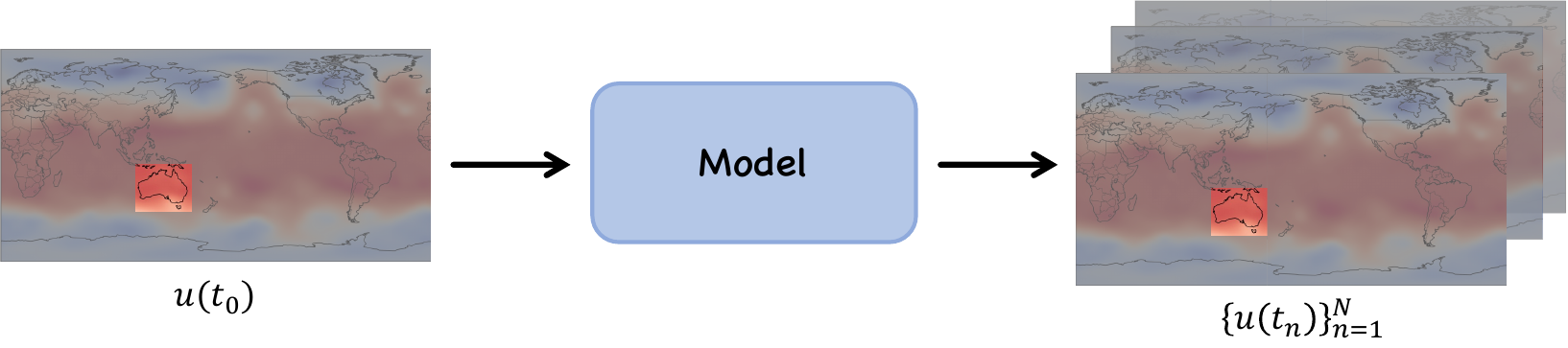}
    \caption{Schematic of the regional forecast for Australia, where only data from the Australian region is used to predict weather conditions within the same area.}
    \label{fig:regional}
\end{figure}

Obtaining global data is often challenging, making it crucial to develop methods that can predict weather using data from specific local regions. As shown in \cref{fig:regional}, we illustrate the forecasting pipeline for regional prediction. We conduct experiments focusing on three regions: North America, South America, and Australia. The data for these regions is extracted as bounding boxes from the 5.625° ERA5 global dataset. \cref{tab:region_statistic} provides the bounding box details for each of the three regions.

\renewcommand{\arraystretch}{1}
\begin{table}[!b]
\centering
\setlength{\tabcolsep}{7pt}

\renewcommand{\arraystretch}{1.35}
\begin{threeparttable}
\begin{tabular}{lccc}
    \toprule
    \textbf{Region} & \textbf{Latitude Range} & \textbf{Longitude Range} & \textbf{Grid Size (lat x lon)} \\
    \midrule
    North America & $(15, 65)$ & $(220, 300)$ & $8 \times 14$ \\
    South America & $(-55, 20)$ & $(270, 330)$ & $14 \times 10$ \\
    Australia & $(-50, 10)$ & $(100, 180)$ & $10 \times 14$ \\
    Global & $(-90, 90)$ & $(0, 360)$ & $32 \times 64$ \\
    \bottomrule
\end{tabular}
\caption{Latitudinal and longitudinal boundaries with grid size for each region.}
\label{tab:region_statistic}
\end{threeparttable}
\end{table}
\renewcommand{\arraystretch}{1}

\newpage 
\section{Full Results}

\begin{table}[!hbt]
    \centering
    \begin{threeparttable}
    \resizebox{\textwidth}{!}{
    \begin{tabular}{lccc cccc}
        \toprule
        & & \multicolumn{3}{c}{$\mathrm{RMSE} \ \downarrow$} &  \multicolumn{3}{c}{$\mathrm{ACC} \ \uparrow$} \\
         \cmidrule(lr){3-5} \cmidrule(lr){6-8}

        Variable & Hours & ClimaX & ClimODE & \NAME & ClimaX & ClimODE & \NAME \\

        \toprule

        \multirow{2}{*}{$z500$} 
        
        & 36 & 126.4 & 259.6 & 159.9 & 1.00 & 0.96 & 0.99 \\
        & 72 & 244.1 & 478.7 & 324.7 & 1.00 & 0.88 & 0.95 \\

        \midrule

       \multirow{2}{*}{$t850$} 
        
        & 36 & 1.25 & 1.75 & 1.22 & 0.97 & 0.94 & 0.97 \\
        & 72 & 1.59 & 2.58 & 1.81 & 0.98 & 0.85 & 0.93 \\

        \midrule

        \multirow{2}{*}{$t2m$} 
        
        & 36 & 1.33 & 1.70 & 1.18 & 0.97 & 0.94 & 0.97 \\
        & 72 & 1.43 & 2.75 & 1.60 & 0.98 & 0.85 & 0.95 \\

        \midrule

        \multirow{2}{*}{$u10$} 
        
        & 36 & 1.57 & 2.25 & 1.57 & 0.93 & 0.83 & 0.93 \\
        & 72 & 2.18 & 3.19 & 2.45 & 0.94 & 0.66 & 0.81 \\

        \midrule

        \multirow{2}{*}{$v10$} 
        
        & 36 & \textcolor{gray}{(N/A)} & 2.29 & 1.61 & \textcolor{gray}{(N/A)} & 0.83 & 0.92 \\
        & 72 & \textcolor{gray}{(N/A)} & 3.30 & 2.50 & \textcolor{gray}{(N/A)} & 0.63 & 0.80 \\

        \bottomrule
    \end{tabular}
    }
    \caption{Latitude-weighted RMSE and ACC for global forecasting at longer lead times.}
    \label{tab:full_long_term}
    \end{threeparttable}
\end{table}

\begin{table}[!hbt]
    \centering
    \begin{threeparttable}
    \resizebox{\textwidth}{!}{
    \begin{tabular}{lccccccc ccccc}
        \toprule
        & & \multicolumn{5}{c}{$\mathrm{RMSE} \ \downarrow$} &  \multicolumn{5}{c}{$\mathrm{ACC} \ \uparrow$} \\
         \cmidrule(lr){3-7} \cmidrule(lr){8-12}

        \scalebox{0.9}{Variable} & \scalebox{0.9}{Hours} & \scalebox{0.8}{wo Source} & \scalebox{0.8}{ViT} & \scalebox{0.8}{DiT} & \scalebox{0.8}{Resnet2D} & \scalebox{0.8}{Resnet3D} & 
        \scalebox{0.8}{wo Source} & \scalebox{0.8}{ViT} & \scalebox{0.8}{DiT} & \scalebox{0.8}{ResNet2D} & \scalebox{0.8}{Resnet3D} \\

        \toprule

        \multirow{4}{*}{$z500$} 
        
        & 6  & 144.6 & 76.0 & 66.4  & 58.6  & 56.3  & 0.99 & 1.00 & 1.00 & 1.00 & 1.00 \\
        & 12 & 164.6 & 94.4 & 81.4  & 75.4  & 73.3  & 0.99 & 1.00 & 1.00 & 1.00 & 1.00 \\
        & 18 & 215.3 & 111.8 & 99.0  & 94.2  & 91.9  & 0.98 & 0.99 & 1.00 & 1.00 & 1.00 \\
        & 24 & 284.3 & 150.0 & 126.8 & 120.0 & 114.5 & 0.96 & 0.99 & 0.99 & 0.99 & 1.00 \\

        \midrule

       \multirow{4}{*}{$t850$} 
        
        & 6  & 1.28 & 0.88 & 0.79 & 0.77 & 0.76 & 0.97 & 0.98 & 0.99 & 0.99 & 0.99 \\
        & 12 & 1.54 & 1.00 & 0.91 & 0.89 & 0.88 & 0.95 & 0.98 & 0.98 & 0.98 & 0.98 \\
        & 18 & 1.58 & 1.08 & 0.99 & 0.97 & 0.95 & 0.95 & 0.98 & 0.98 & 0.98 & 0.98 \\
        & 24 & 1.87 & 1.22 & 1.08 & 1.06 & 1.04 & 0.93 & 0.97 & 0.98 & 0.98 & 0.98 \\

        \midrule

        \multirow{4}{*}{$t2m$} 
        
        & 6  & 2.55 & 0.88 & 0.79 & 0.79 & 0.78 & 0.87 & 0.99 & 0.99 & 0.99 & 0.99 \\
        & 12 & 2.89 & 1.02 & 0.92 & 0.91 & 0.89 & 0.83 & 0.98 & 0.98 & 0.98 & 0.98 \\
        & 18 & 2.11 & 1.06 & 0.98 & 0.97 & 0.95 & 0.91 & 0.98 & 0.98 & 0.98 & 0.98 \\
        & 24 & 2.65 & 1.15 & 1.02 & 1.02 & 0.98 & 0.86 & 0.97 & 0.98 & 0.98 & 0.98 \\

        \midrule

        \multirow{4}{*}{$u10$} 
        
        & 6  & 1.59 & 0.96 & 0.90 & 0.88 & 0.88 & 0.92 & 0.97 & 0.98 & 0.98 & 0.98 \\
        & 12 & 1.73 & 1.08 & 1.02 & 1.01 & 1.00 & 0.91 & 0.96 & 0.97 & 0.97 & 0.97 \\
        & 18 & 1.85 & 1.21 & 1.14 & 1.13 & 1.13 & 0.89 & 0.95 & 0.96 & 0.96 & 0.96 \\
        & 24 & 2.31 & 1.38 & 1.28 & 1.27 & 1.26 & 0.84 & 0.94 & 0.95 & 0.95 & 0.95 \\

        \midrule

        \multirow{4}{*}{$v10$} 
        
        & 6  & 1.63 & 0.99 & 0.93 & 0.91 & 0.90 & 0.92 & 0.97 & 0.97 & 0.98 & 0.98 \\
        & 12 & 1.78 & 1.12 & 1.06 & 1.04 & 1.04 & 0.90 & 0.97 & 0.97 & 0.97 & 0.97 \\
        & 18 & 1.93 & 1.24 & 1.18 & 1.16 & 1.16 & 0.89 & 0.96 & 0.96 & 0.96 & 0.96 \\
        & 24 & 2.40 & 1.41 & 1.32 & 1.30 & 1.29 & 0.82 & 0.94 & 0.95 & 0.95 & 0.95 \\

        \bottomrule
    \end{tabular}}
    \caption{Full results on source model architectures shown in \cref{fig:ablation_source_net}.}
    \label{tab:ablation_source}
    \end{threeparttable}
\end{table}

\begin{table}[!hbt]
    \centering
    \begin{threeparttable}
    \resizebox{\textwidth}{!}{
    \begin{tabular}{lccccccc ccccc}
        \toprule
        & & \multicolumn{5}{c}{$\mathrm{RMSE} \ \downarrow$} &  \multicolumn{5}{c}{$\mathrm{ACC} \ \uparrow$} \\
         \cmidrule(lr){3-7} \cmidrule(lr){8-12}

        \scalebox{0.9}{Variable} & \scalebox{0.9}{Hours} & \scalebox{0.8}{$f_v(\frac{\Delta u}{\Delta t})$} & \scalebox{0.8}{$f_v(u)$} & \scalebox{0.8}{$f_v(\nabla u)$} & \scalebox{0.8}{$f_v(u,\nabla u, \frac{\Delta u}{\Delta t})$} & \scalebox{0.8}{$f_v(u,\nabla u)$} & 
        \scalebox{0.8}{$f_v(\frac{\Delta u}{\Delta t})$} & \scalebox{0.8}{$f_v(u)$} & \scalebox{0.8}{$f_v(\nabla u)$} & \scalebox{0.8}{$f_v(u,\nabla u, \frac{\Delta u}{\Delta t})$} & \scalebox{0.8}{$f_v(u,\nabla u)$} \\

        \toprule

        \multirow{4}{*}{$z500$} 
        
        & 6  & 71.0  & 73.6  & 59.8  & 61.4  & 56.3  & 1.00 & 1.00 & 1.00 & 1.00 & 1.00 \\
        & 12 & 100.6 & 101.2 & 79.4  & 83.5  & 73.3  & 0.99 & 1.00 & 1.00 & 1.00 & 1.00 \\
        & 18 & 134.0 & 129.7 & 101.6 & 108.1 & 91.9  & 0.99 & 0.99 & 1.00 & 0.99 & 1.00 \\
        & 24 & 172.8 & 162.6 & 128.5 & 137.6 & 114.5 & 0.98 & 0.99 & 0.99 & 0.99 & 1.00 \\

        \midrule

       \multirow{4}{*}{$t850$} 
        
        & 6  & 0.83 & 0.83 & 0.77 & 0.79 & 0.76 & 0.99 & 0.99 & 0.99 & 0.99 & 0.99 \\
        & 12 & 1.00 & 0.98 & 0.90 & 0.92 & 0.88 & 0.98 & 0.98 & 0.98 & 0.98 & 0.98 \\
        & 18 & 1.12 & 1.09 & 0.99 & 1.02 & 0.95 & 0.97 & 0.98 & 0.98 & 0.98 & 0.98 \\
        & 24 & 1.25 & 1.20 & 1.08 & 1.12 & 1.04 & 0.97 & 0.97 & 0.98 & 0.97 & 0.98 \\

        \midrule

        \multirow{4}{*}{$t2m$} 
        
        & 6  & 0.91 & 0.89 & 0.80 & 0.82 & 0.78 & 0.98 & 0.98 & 0.99 & 0.99 & 0.99 \\
        & 12 & 1.09 & 1.04 & 0.93 & 0.95 & 0.89 & 0.98 & 0.98 & 0.98 & 0.98 & 0.98 \\
        & 18 & 1.13 & 1.10 & 0.99 & 1.02 & 0.95 & 0.98 & 0.98 & 0.98 & 0.98 & 0.98 \\
        & 24 & 1.16 & 1.13 & 1.03 & 1.06 & 0.98 & 0.97 & 0.98 & 0.98 & 0.98 & 0.98 \\

        \midrule

        \multirow{4}{*}{$u10$} 
        
        & 6  & 0.96 & 0.95 & 0.89 & 0.90 & 0.88 & 0.97 & 0.97 & 0.98 & 0.98 & 0.98 \\
        & 12 & 1.13 & 1.12 & 1.03 & 1.04 & 1.00 & 0.96 & 0.96 & 0.97 & 0.97 & 0.97 \\
        & 18 & 1.30 & 1.28 & 1.15 & 1.18 & 1.13 & 0.95 & 0.95 & 0.96 & 0.96 & 0.96 \\
        & 24 & 1.50 & 1.45 & 1.30 & 1.33 & 1.26 & 0.93 & 0.94 & 0.95 & 0.95 & 0.95 \\

        \midrule

        \multirow{4}{*}{$v10$} 
        
        & 6  & 0.98 & 0.98 & 0.92 & 0.93 & 0.90 & 0.97 & 0.97 & 0.97 & 0.97 & 0.98 \\
        & 12 & 1.16 & 1.16 & 1.06 & 1.07 & 1.04 & 0.96 & 0.96 & 0.97 & 0.97 & 0.97 \\
        & 18 & 1.34 & 1.32 & 1.19 & 1.21 & 1.16 & 0.95 & 0.95 & 0.96 & 0.96 & 0.96 \\
        & 24 & 1.54 & 1.50 & 1.33 & 1.37 & 1.29 & 0.93 & 0.93 & 0.95 & 0.94 & 0.95 \\

        \bottomrule
    \end{tabular}}
    \caption{Full results of different input configurations of the velocity model in \cref{fig:ablation_v_net}.}
    \label{tab:ablation_velocity}
    \end{threeparttable}
\end{table}

\setlength{\heavyrulewidth}{1pt}
\begin{table}[!hbt]
    \centering
    \begin{threeparttable}
    \resizebox{\textwidth}{!}{
    \begin{tabular}{lccccccc ccccc}
        \toprule
        & & \multicolumn{5}{c}{$\mathrm{RMSE} \ \downarrow$} &  \multicolumn{5}{c}{$\mathrm{ACC} \ \uparrow$} \\
         \cmidrule(lr){3-7} \cmidrule(lr){8-12}

        \scalebox{0.9}{Variable} & \scalebox{0.9}{Hours} & \scalebox{0.8}{$\Delta t=1$} & \scalebox{0.8}{$\Delta t=2$} & \scalebox{0.8}{$\Delta t=3$} & \scalebox{0.8}{$\Delta t=12$} & \scalebox{0.8}{$\Delta t=24$} & 
        \scalebox{0.8}{$\Delta t=1$} & \scalebox{0.8}{$\Delta t=2$} & \scalebox{0.8}{$\Delta t=3$} & \scalebox{0.8}{$\Delta t=12$} & \scalebox{0.8}{$\Delta t=24$} \\

        \toprule

        \multirow{4}{*}{$z500$} 
        
        & 6  & 71.0  & 86.8  & 88.8  & 107.8 & 140.5 & 1.00 & 1.00 & 1.00 & 0.99 & 0.98 \\
        & 12 & 100.6 & 118.5 & 128.7 & 164.3 & \textcolor{gray}{NaN} & 0.99 & 0.99 & 0.99 & 0.98 & \textcolor{gray}{NaN} \\
        & 18 & 134.0 & 157.3 & 161.3 & \textcolor{gray}{NaN} & \textcolor{gray}{NaN} & 0.99 & 0.99 & 0.99 & \textcolor{gray}{NaN} & \textcolor{gray}{NaN} \\
        & 24 & 172.8 & \textcolor{gray}{NaN}   & \textcolor{gray}{NaN}   & \textcolor{gray}{NaN} & \textcolor{gray}{NaN} & 0.98 & \textcolor{gray}{NaN}  & \textcolor{gray}{NaN}  & \textcolor{gray}{NaN} & \textcolor{gray}{NaN} \\

        \midrule

       \multirow{4}{*}{$t850$} 
        
        & 6  & 0.83 & 0.89 & 0.91 & 0.95 & 1.04 & 0.99 & 0.98 & 0.98 & 0.97 & 0.97 \\
        & 12 & 1.00 & 1.10 & 1.11 & 1.18 & \textcolor{gray}{NaN} & 0.98 & 0.98 & 0.97 & 0.97 & \textcolor{gray}{NaN} \\
        & 18 & 1.12 & 1.24 & 1.25 & \textcolor{gray}{NaN} & \textcolor{gray}{NaN} & 0.97 & 0.97 & 0.97 & \textcolor{gray}{NaN} & \textcolor{gray}{NaN} \\
        & 24 & 1.25 & \textcolor{gray}{NaN}  & \textcolor{gray}{NaN}  & \textcolor{gray}{NaN} & \textcolor{gray}{NaN} & 0.97 & \textcolor{gray}{NaN}  & \textcolor{gray}{NaN} & \textcolor{gray}{NaN} & \textcolor{gray}{NaN} \\

        \midrule

        \multirow{4}{*}{$t2m$} 
        
        & 6  & 0.91 & 0.98 & 1.00 & 1.12 & 1.35 & 0.98 & 0.98 & 0.98 & 0.97 & 0.96 \\
        & 12 & 1.09 & 1.18 & 1.23 & 1.35 & \textcolor{gray}{NaN} & 0.98 & 0.97 & 0.97 & 0.96 & \textcolor{gray}{NaN} \\
        & 18 & 1.13 & 1.21 & 1.28 & \textcolor{gray}{NaN} & \textcolor{gray}{NaN} & 0.98 & 0.97 & 0.97 & \textcolor{gray}{NaN} & \textcolor{gray}{NaN} \\
        & 24 & 1.16 & \textcolor{gray}{NaN}  & \textcolor{gray}{NaN}  & \textcolor{gray}{NaN} & \textcolor{gray}{NaN} & 0.97 & \textcolor{gray}{NaN}  & \textcolor{gray}{NaN}  & \textcolor{gray}{NaN} & \textcolor{gray}{NaN} \\

        \midrule

        \multirow{4}{*}{$u10$} 
        
        & 6  & 0.96 & 1.02 & 1.03 & 1.07 & 1.11 & 0.97 & 0.97 & 0.97 & 0.97 & 0.96 \\
        & 12 & 1.13 & 1.24 & 1.23 & 1.41 & \textcolor{gray}{NaN} & 0.96 & 0.95 & 0.95 & 0.93 & \textcolor{gray}{NaN} \\
        & 18 & 1.30 & 1.44 & 1.42 & \textcolor{gray}{NaN} & \textcolor{gray}{NaN} & 0.95 & 0.93 & 0.94 & \textcolor{gray}{NaN} & \textcolor{gray}{NaN} \\
        & 24 & 1.50 & \textcolor{gray}{NaN}  & \textcolor{gray}{NaN}  & \textcolor{gray}{NaN} & \textcolor{gray}{NaN} & 0.93 & \textcolor{gray}{NaN}  & \textcolor{gray}{NaN}  & \textcolor{gray}{NaN} & \textcolor{gray}{NaN} \\

        \midrule

        \multirow{4}{*}{$v10$} 
        
        & 6  & 0.98 & 1.06 & 1.07 & 1.11 & 1.16 & 0.97 & 0.97 & 0.96 & 0.96 & 0.96 \\
        & 12 & 1.16 & 1.28 & 1.31 & 1.44 & \textcolor{gray}{NaN} & 0.96 & 0.95 & 0.94 & 0.94 & \textcolor{gray}{NaN} \\
        & 18 & 1.34 & 1.47 & 1.52 & \textcolor{gray}{NaN} & \textcolor{gray}{NaN} & 0.95 & 0.93 & 0.93 & \textcolor{gray}{NaN} & \textcolor{gray}{NaN} \\
        & 24 & 1.54 & \textcolor{gray}{NaN}  & \textcolor{gray}{NaN}  & \textcolor{gray}{NaN} & \textcolor{gray}{NaN} & 0.93 & \textcolor{gray}{NaN}  & \textcolor{gray}{NaN}  & \textcolor{gray}{NaN} & \textcolor{gray}{NaN} \\

        \bottomrule
    \end{tabular}}
    \caption{Full result of the time interval $\Delta t$ for estimating $\frac{\Delta u}{\Delta t}$ in \cref{fig:intro}b. \textcolor{gray}{NaN} indicates that numerical instability occurred during ODE inference.}
    \label{tab:ablation_time_interval}
    \end{threeparttable}
\end{table}

\clearpage
\section{Visualization}
\begin{figure}[htb]
    \centering
    \includegraphics[width=\linewidth]{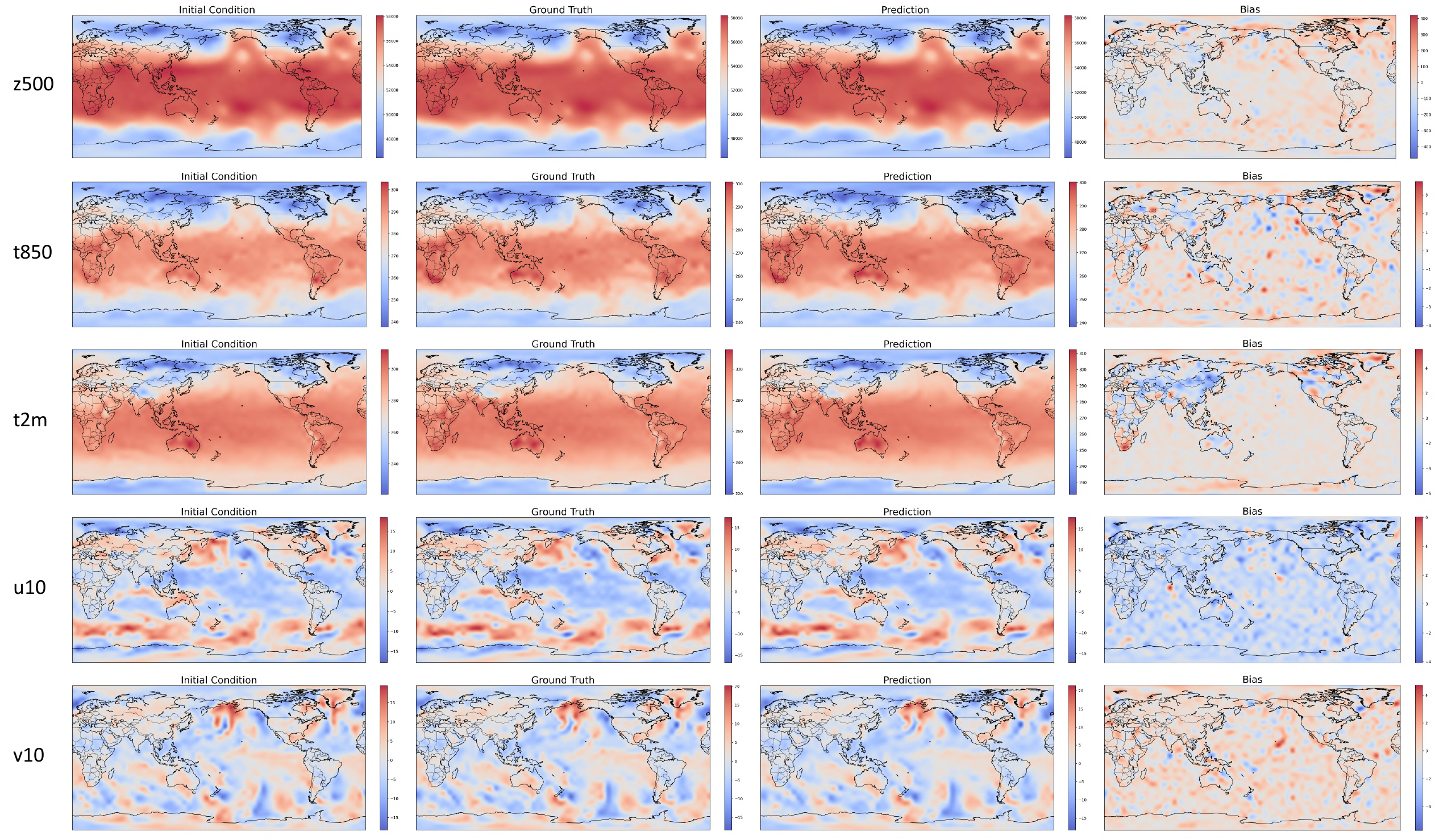}
    \caption{Example 6-hour lead time forecasts from \NAME compared to ground truth ERA5 data.}
    \label{fig:predict_6h_vis}
\end{figure}

\begin{figure}[htb]
    \centering
    \includegraphics[width=\linewidth]{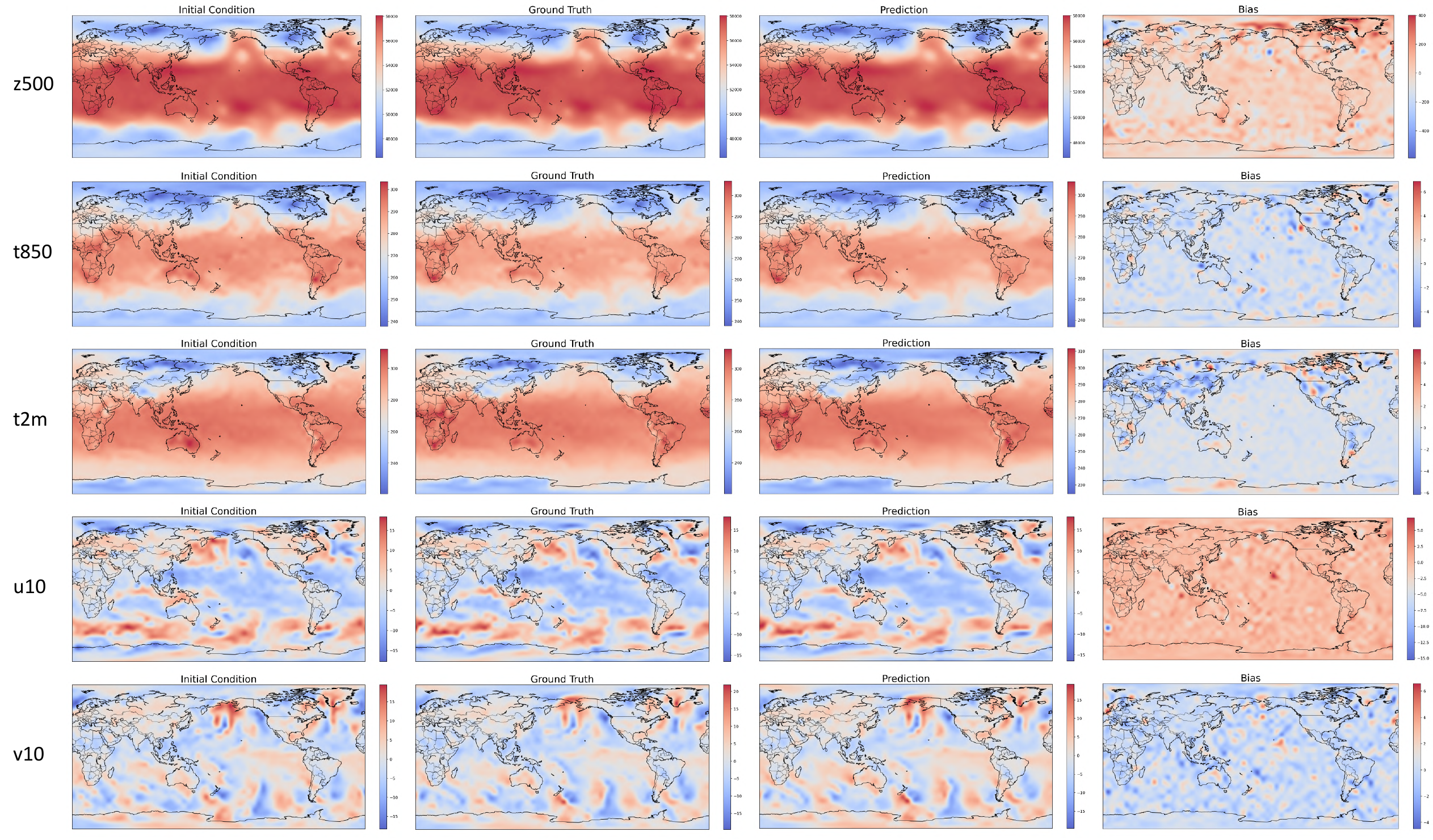}
    \caption{Example 12-hour lead time forecasts from \NAME compared to ground truth ERA5 data.}
    \label{fig:predict_12h_vis}
\end{figure}

\begin{figure}[htb]
    \centering
    \includegraphics[width=\linewidth]{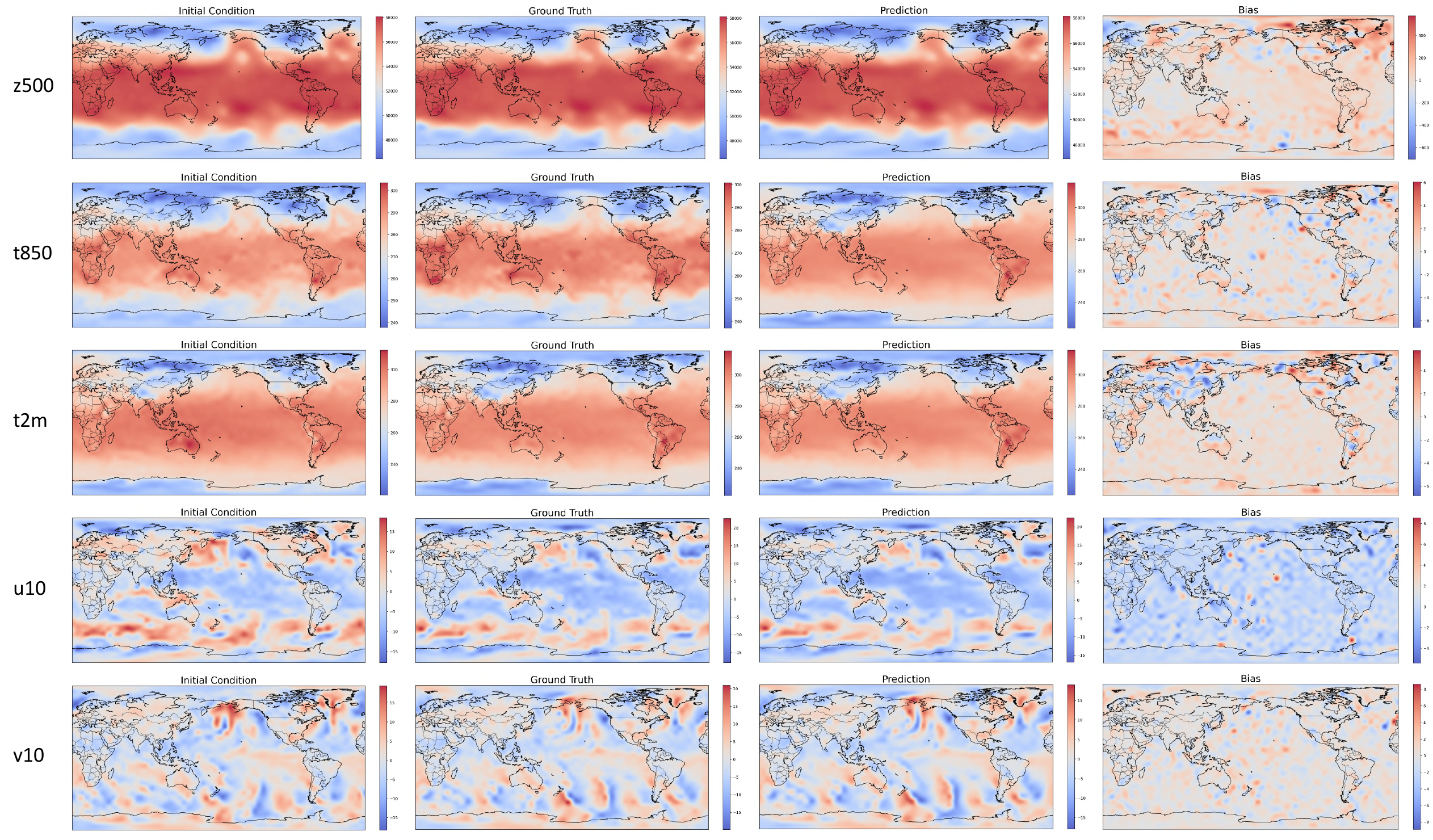}
    \caption{Example 18-hour lead time forecasts from \NAME compared to ground truth ERA5 data.}
    \label{fig:predict_18h_vis}
\end{figure}

\begin{figure}[htb]
    \centering
    \includegraphics[width=\linewidth]{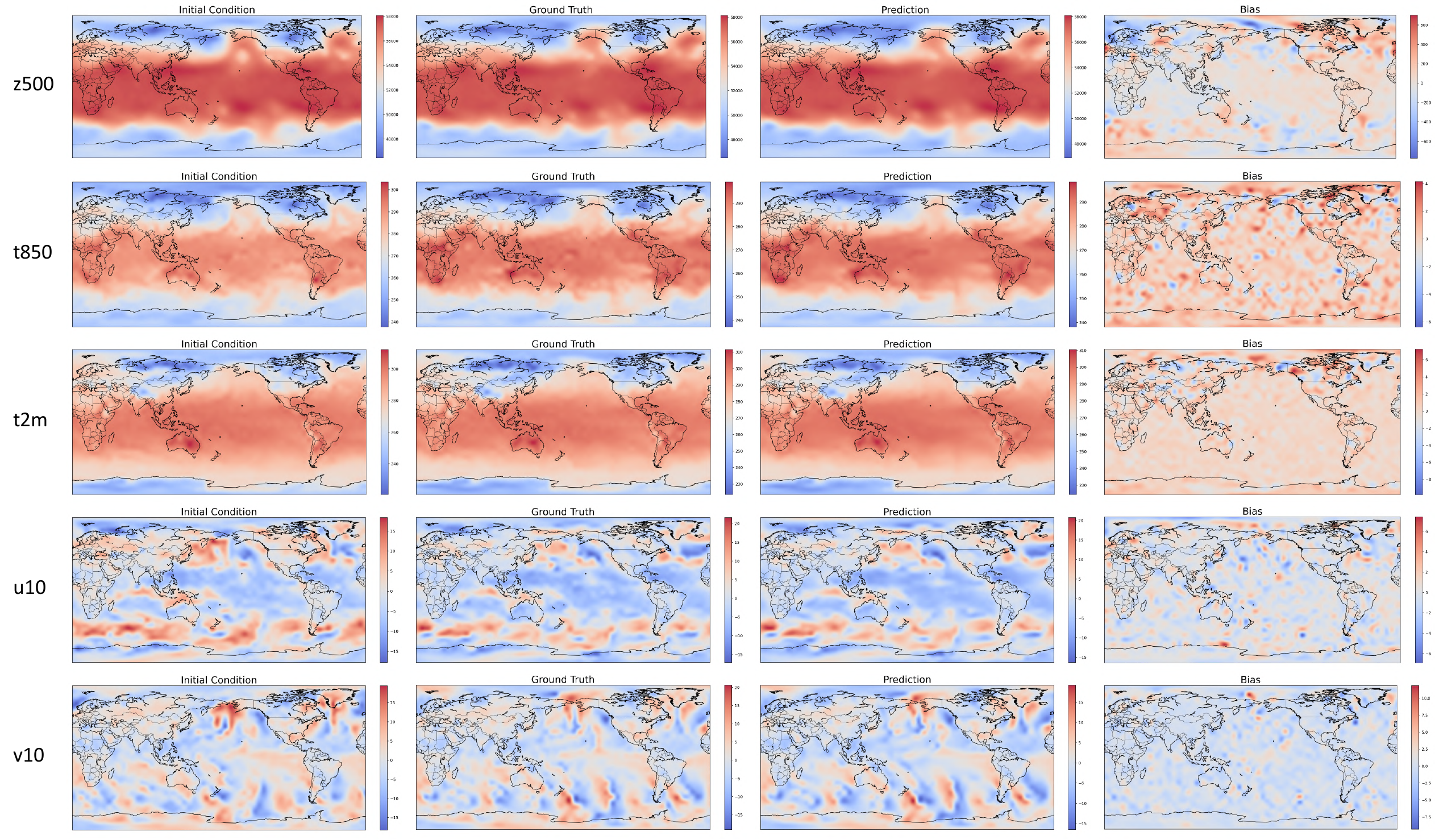}
    \caption{Example 24-hour lead time forecasts from \NAME compared to ground truth ERA5 data.}
    \label{fig:predict_24h_vis}
\end{figure}

\cref{fig:predict_6h_vis} to \cref{fig:predict_24h_vis} provide visual comparisons between \NAME's forecasts and the ground truth ERA5 data at different lead times (6h, 12h, 18h, and 24h). \cref{fig:predict_24h_compare} illustrates the output from the advection ODE and the source model, demonstrating that the advection ODE captures global features, while the source model captures local features.

\afterpage{
\begin{figure}[htb]
    \centering
    \includegraphics[width=\linewidth]{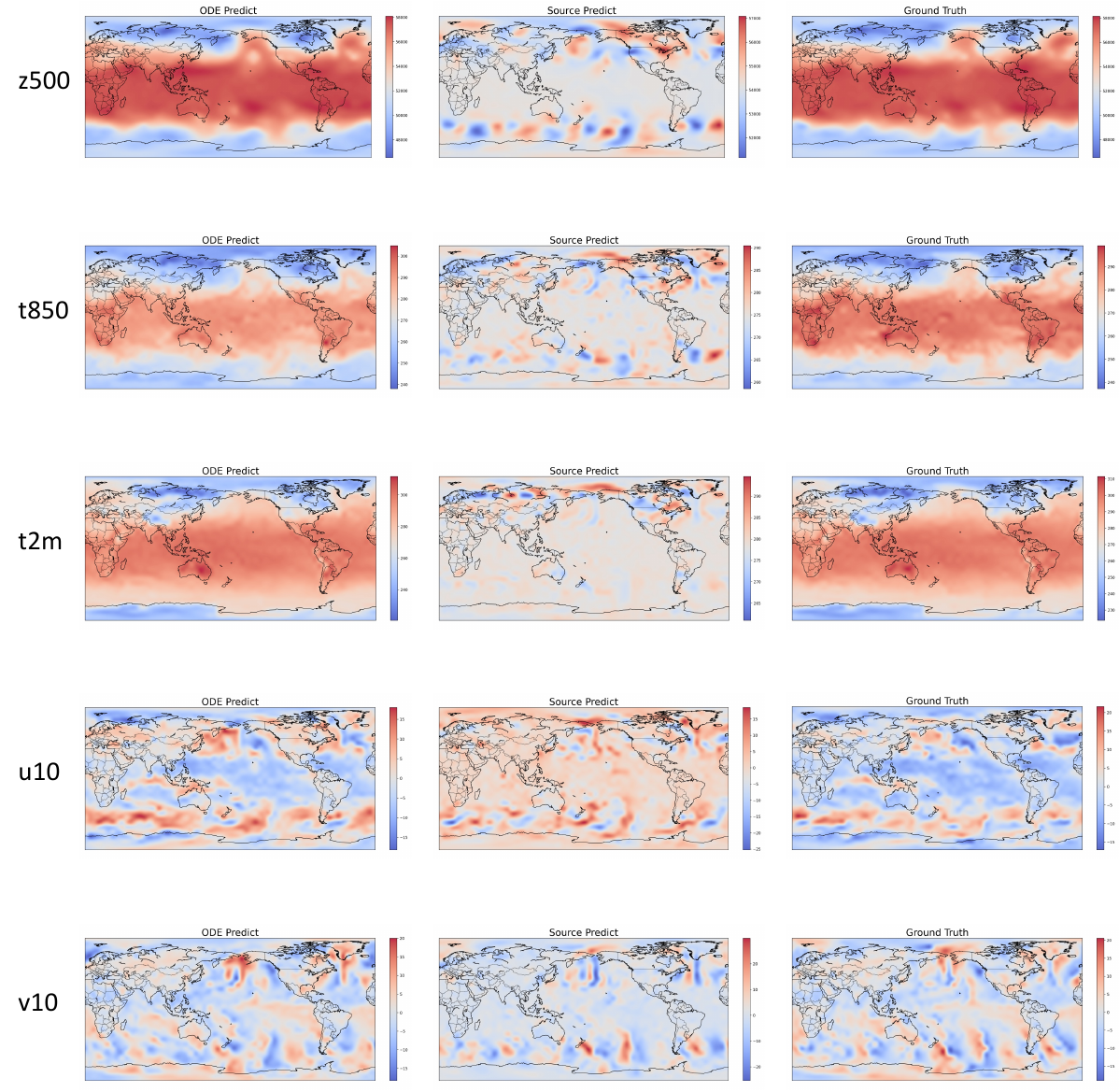}
    \caption{Example 24-hour lead time forecasts from the advection ODE, source model, and ground truth comparison.}
    \label{fig:predict_24h_compare}
\end{figure}
}

\end{document}